\documentclass[sigconf]{acmart}

\usepackage{graphicx}

\usepackage{xcolor}

\usepackage{algorithm}
\usepackage[noend]{algpseudocode}

\newcommand*\Input[1]{\Statex \textbf{Input:} #1}
\newcommand*\Output[1]{\Statex \textbf{Output:} #1}
\algrenewcommand\alglinenumber[1]{#1}

\usepackage{multirow}

\usepackage{enumitem}
\setlist[itemize]{leftmargin=*, topsep=0pt, itemsep=0pt, parsep=0pt, partopsep=0pt}
\setlist[enumerate]{leftmargin=*, topsep=0pt, itemsep=0pt, parsep=0pt, partopsep=0pt}

\copyrightyear{2022} 
\acmYear{2022} 
\setcopyright{acmlicensed}\acmConference[KDD '22]{Proceedings of the 28th ACM SIGKDD Conference on Knowledge Discovery and Data Mining}{August 14--18, 2022}{Washington, DC, USA}
\acmBooktitle{Proceedings of the 28th ACM SIGKDD Conference on Knowledge Discovery and Data Mining (KDD '22), August 14--18, 2022, Washington, DC, USA}
\acmPrice{15.00}
\acmDOI{10.1145/3534678.3539068}
\acmISBN{978-1-4503-9385-0/22/08}

\begin{document}

\title{Embedding Compression with Hashing for Efficient Representation Learning in Large-Scale Graph}

\author{Chin-Chia Michael Yeh}
\email{miyeh@visa.com}
\affiliation{%
  \institution{Visa Research}
}
\author{Mengting Gu}
\email{mengu@visa.com}
\affiliation{%
  \institution{Visa Research}
}
\author{Yan Zheng}
\email{yazheng@visa.com}
\affiliation{%
  \institution{Visa Research}
}
\author{Huiyuan Chen}
\email{hchen@visa.com}
\affiliation{%
  \institution{Visa Research}
}
\author{Javid Ebrahimi}
\email{jebrahim@visa.com}
\affiliation{%
  \institution{Visa Research}
}
\author{Zhongfang Zhuang}
\email{zzhuang@visa.com}
\affiliation{%
  \institution{Visa Research}
}
\author{Junpeng Wang}
\email{junpenwa@visa.com}
\affiliation{%
  \institution{Visa Research}
}
\author{Liang Wang}
\email{liawang@visa.com}
\affiliation{%
  \institution{Visa Research}
}
\author{Wei Zhang}
\email{wzhan@visa.com}
\affiliation{%
  \institution{Visa Research}
}

\renewcommand{\shortauthors}{Chin-Chia Michael Yeh et al.} 

\begin{abstract}
Graph neural networks (GNNs) are deep learning models designed specifically for graph data, and they typically rely on node features as the input to the first layer. 
When applying such a type of network on the graph without node features, one can extract simple graph-based node features (e.g., number of degrees) or learn the input node representations (i.e., embeddings) when training the network. 
While the latter approach, which trains node embeddings, more likely leads to better performance, the number of parameters associated with the embeddings grows linearly with the number of nodes. 
It is therefore impractical to train the input node embeddings together with GNNs within graphics processing unit (GPU) memory in an end-to-end fashion when dealing with industrial-scale graph data. 
Inspired by the embedding compression methods developed for natural language processing (NLP) tasks, we develop a node embedding compression method where each node is compactly represented with a bit vector instead of a floating-point vector. 
The parameters utilized in the compression method can be trained together with GNNs.
We show that the proposed node embedding compression method achieves superior performance compared to the alternatives.
\end{abstract}



\maketitle

\section{Introduction}
Graph neural networks (GNNs) are representation learning methods for graph data.
They learn the node representation from input node features~$\mathbf{X}$ and its graph~$\mathcal{G}$ where the node features~$\mathbf{X}$ are used as the input node representations to the first layer of the GNN and the graph~$\mathcal{G}$ dictates the propagation of information~\cite{kipf2016semi,hamilton2017inductive,zhou2020graph}.
However, the input node features~$\mathbf{X}$ may not always be available for certain datasets.
To apply GNNs on a graph without node features~$\mathbf{X}$, we could either 1) extract simple graph-based node features (e.g., number of degrees) from the graph~$\mathcal{G}$ or 2) use embedding learning methods to learn the node embeddings as features~$\mathbf{X}$~\cite{duong2019node}.
While both approaches are valid, it has been shown that the second approach constantly outperforms the first one with a noticeable margin~\cite{duong2019node}, and most recent methods learn the node embeddings jointly with the parameters of GNNs~\cite{he2017neural,he2020lightgcn,wang2019neural}.

Learning node features (or embedding)~$\mathbf{X}$ for a small graph can be easily conducted.
But, as the size of the embedding matrix~$\mathbf{X}$ grows linearly with the number of nodes, scalability quickly becomes a problem, especially when attempting to apply such a method to industrial-grade graph data.
For example, there are more than 1 billion Visa cards.
If 1 billion of these cards are modeled as nodes in a graph, the memory cost for the embedding layer alone will be 238 gigabytes for 64-dimensional single-precision floating-point embeddings.
Such memory cost is beyond the capability of common graphics processing unit (GPU).
To solve the scalability issue, we adopt the embedding compression idea originally developed for natural language processing (NLP) tasks~\cite{suzuki2016learning,shu2017compressing,svenstrup2017hash,takase2020all}.

\begin{figure*}[ht]
\begin{center}
\includegraphics[width=0.75\linewidth]{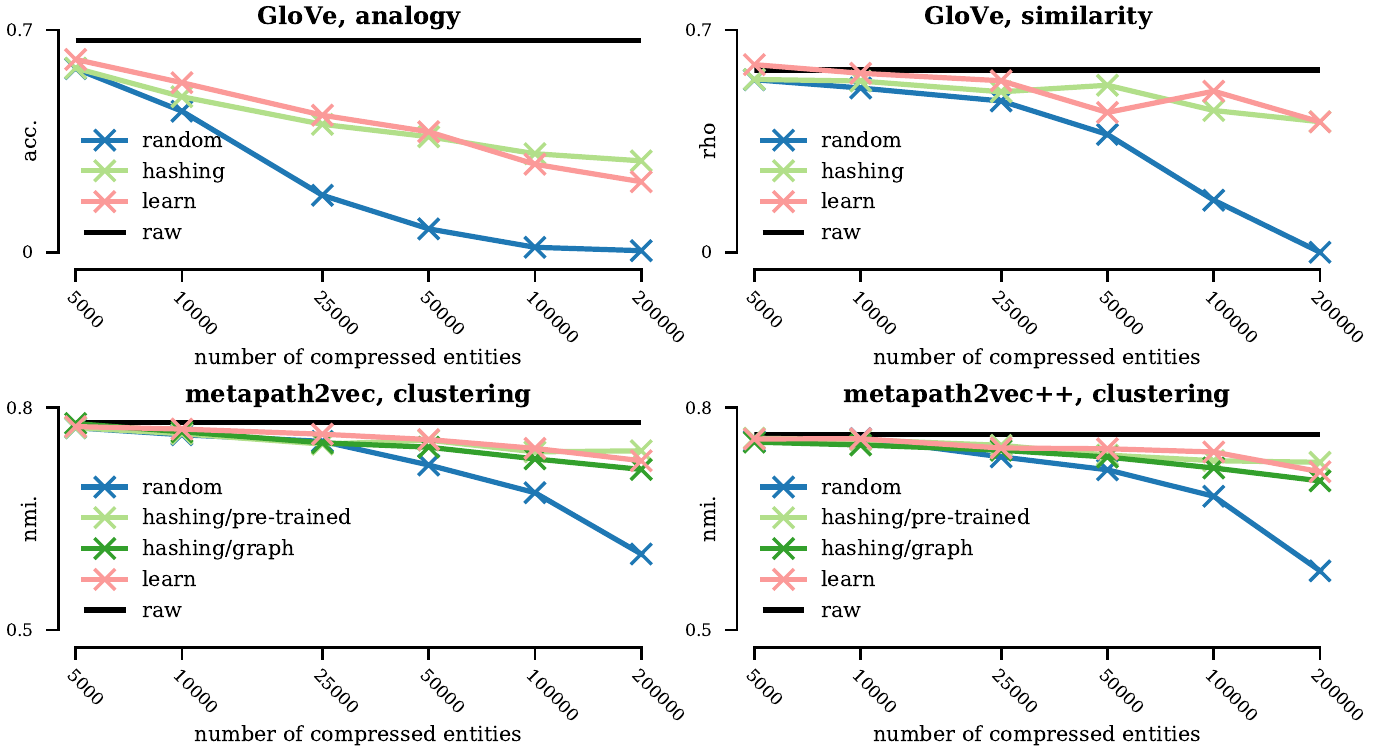}
\end{center}
\caption{
Three coding schemes are tested: 1) random coding/\texttt{ALONE}, 2) hashing-based coding/the proposed method, and 3) learning-based coding scheme with an autoencoder.
For \texttt{GloVe} embeddings, we apply the hashing-based coding method on the pre-trained embeddings.
For \texttt{metapath2vec} and \texttt{metapath2vec++} embeddings, we apply the hashing-based coding method on either the pre-trained embeddings or the adjacency matrix from the graph.
The horizontal line labeled with ``raw" shows the performance of the original embeddings' performance without any compression.
The y-axis of each sub-figure is the performance measurement (the higher the better).
See Section~\ref{sec:pretrain} for more details.
}
\label{fig:pretrain}
\end{figure*}

Particularly, we study the \texttt{ALONE} method proposed by~\cite{takase2020all}.
\texttt{ALONE} represents each word using a randomly generated compositional code vector~\cite{takase2020all}, which significantly reduces the memory cost; then a decoder model uncompresses the compositional code vector into a floating-point vector.
The bit size of the compositional code vector is parametrized by $c$ and $m$ where $c$ is the cardinality of each element in the code vector and $m$ is the length of the code vector.
For example, if we set $c=4$ and $m=6$, one valid code vector is $\mathtt{\left[2, 0, 3, 1, 0, 1\right]}$ where the length of the vector is 6 and each element in the vector is within the set $\mathtt{\{0, 1, 2, 3\}}$.
The code vector can be converted to a bit vector of length $m \log_2 c$ by representing each element in the code vector as a binary number and concatenating the resulting binary numbers\footnotemark.
Continuing the example, the code vector $\mathtt{\left[2, 0, 3, 1, 0, 1\right]}$ can be compactly stored as $\mathtt{\left[10\,00\,11\,01\,00\,01\right]}$.
\footnotetext{
The conversion mechanism is more space-efficient when $c$ is set to a power of 2.}
Using the conversion trick, it only requires 48 bits to store each word with the parametrization ($c=64, m=8$, $8 \log_2 64=48$ bits) used by~\cite{takase2020all} in their experiments.

The coding scheme can uniquely represent up to $2^{48}$ word (or sub-word) tokens, which is way beyond the number of tokens used in conventional NLP models~\cite{vaswani2017attention,takase2019positional}.
However, generating the code vectors in a random fashion hinders the model performance.
One way to quickly benchmark different embedding compression methods is by evaluating the performance of reconstructed (or \textit{uncompressed}) embeddings.
As shown in Figure~\ref{fig:pretrain}, when a model compresses more embeddings, the performance of the uncompressed embeddings drops considerably when \texttt{ALONE}~\cite{takase2020all} is used (see lines labeled as ``random").
The phenomenon is observed in our experiments with \texttt{GloVe} word embeddings~\cite{pennington2014glove} on both word analogy/similarity tasks and \texttt{metapath2vec}/\texttt{metapath2vec++} node embeddings~\cite{dong2017metapath2vec} on node clustering task.
Note, the experiments presented in Figure~\ref{fig:pretrain} (i.e., reconstruction experiments) are only proxies to the real use scenarios (i.e., node classification and link prediction with GNNs).
In the intended use scenarios (Section~\ref{sec:gnn_embedding} and Section~\ref{sec:mci_embedding}), we do not use any pre-trained embeddings.

The possible root cause of the performance degradation is the need of a more expensive decoder to model the larger variance (from the increasing number of randomly generated code vectors).
In order to solve this problem, we replace the random vector generation part of \texttt{ALION} with an efficient random projection hashing algorithm, which better leverages the auxiliary information of the graph~$\mathcal{G}$, such as its adjacency matrix.
The adopted hashing method is locality-sensitive hashing (LSH)~\cite{charikar2002similarity}, as it hashes entities with similar auxiliary information into similar code vectors.
The auxiliary information help us reduce the variance in the code vectors (compared to that of the randomly generated code vectors), which eliminates the need of an expensive decoder.
In Figure~\ref{fig:pretrain}, our proposed hashing-based coding (see lines labeled as ``hashing") outperformed the random coding in all scenarios.
Similar to \texttt{ALONE}, the proposed method does not introduce additional training stages.
On top of that, the memory footprint is identical to \texttt{ALONE} as the proposed method only replaces the coding scheme. 

\sloppy
In addition to the proxy tasks of pre-trained embedding reconstruction, we also compare the effectiveness of different coding schemes where the GNN models and the decoder model are trained together in an end-to-end fashion.
Particularly, we trained four different GNN models~\cite{kipf2016semi,hamilton2017inductive,xu2018powerful,wu2019simplifying} on five different node classification/link prediction datasets as well as our in-house large scale transaction dataset.
The experiment results have confirmed the superb performance of the proposed hashing-based coding scheme. 
To sum up, our contributions include:
\begin{itemize}
    \item We propose a novel hashing-based coding scheme for large-scale graphs, which is compatible with most of the GNNs and achieves superior performance compared to existing methods.
    \item We show how the improved embedding compression method can be applied to GNNs \textit{without} any pre-training.
    \item We confirm the effectiveness of our embedding compression method in \textit{proxy} pre-trained embedding reconstruction tasks, node classification/link prediction tasks with GNNs, and an industrial-scale merchant category identification task.
\end{itemize}

\section{Related Work}
The embedding compression problem is extensively studied for NLP tasks because of the memory cost associated with storing the embedding vectors.
One of the most popular strategies is parameter sharing~\cite{suzuki2016learning,shu2017compressing,svenstrup2017hash,takase2020all}.
For example, Suzuki and Nagata~\cite{suzuki2016learning} train a small set of sub-vectors shared by all words called ``reference vectors" where each word embedding is constructed by concatenating different sub-vectors together.
Both the shared sub-vectors and sub-vector assignments are optimized during the training time.
The resulting compressed representation is capable of representing each word compactly, but the training process is memory costly as it still needs to train the full embedding matrix to solve the sub-vector assignment problem.
Therefore, the method proposed by~\cite{suzuki2016learning} is not suitable for representing a large set of entities.

Shu and Nakayama~\cite{shu2017compressing} train an encoder-decoder model (i.e., autoencoder) where the encoder converts a pre-trained embedding into the corresponding compositional code representation, while the decoder reconstructs the pre-trained embedding from the compositional code.
Once the encoder-decoder is trained, all the pre-trained embeddings are converted to the compact compositional code representation using the encoder; then the decoder can be trained together with the downstream models.
Because the memory cost associated with the compositional code is much smaller than the raw embeddings and the decoder is shared by all words, the method reduces the overall memory consumption associated with representing words.
However, since it also requires training the embeddings before training the encoder-decoder, the training process still has high memory cost associated with the conventional embedding training similar to the previous work.
In other words, the method presented in~\cite{shu2017compressing} is not applicable to compress a large set of entities either.

Svenstrup et al.~\cite{svenstrup2017hash} represents each word compactly with a unique integer and $k$ floating-point values where $k$ is much smaller than the dimension of the embedding.
To obtain the embedding from the compact representation of a word, $k$ hash functions\footnotemark are used to hash the word's associated unique integer to an integer in $[0, c)$ where $c$ is much smaller than the number of words.
\footnotetext{
The hash function used in~\cite{svenstrup2017hash} is the hash function proposed by~\cite{carter1979universal} for hashing integers.}
Next, $k$ vectors are extracted from a set of $c$ learnable ``component vectors" based on the output of the hash function.
The final embedding of the word is generated by computing the weighted sum of the $k$ vectors where the weights are based on the $k$ learnable floating-point values associated with the word.
Similar to our work, Svenstrup et al.~\cite{svenstrup2017hash} also uses hash functions in their proposed method.
But, the role of the hash function is different: Svenstrup et al.~\cite{svenstrup2017hash} uses hash functions for reducing cardinality while we use hash functions to perform LSH.
On top of that, as the $k$ learnable floating-point values are associated with each word in~\cite{svenstrup2017hash}, their method has its parameter size grown linearly with respect to the vocabulary size, which makes the method not ideal for our application.

The \texttt{ALONE} method proposed by~\cite{takase2020all} represents each word with a randomly generated compositional code, and the embedding is obtained by inputting the compositional code to a decoder where the number of learnable parameters in the decoder model is independent of the vocabulary size.
The \texttt{ALONE} method satisfies all the requirements for our application; however, the performance suffers when the vocabulary size increases compared to the autoencoder-based approach~\cite{shu2017compressing} as demonstrated in Figure~\ref{fig:pretrain} (see ``random" versus ``learn").
In contrast, our proposed method has similar performance compared to the autoencoder-based approach~\cite{shu2017compressing} but does not require additional training phases.

Learning-to-hash methods are another set of methods that are concerned with the compression of data~\cite{tan2020learning,qin2020ghashing}.
When learning-to-hash methods are applied to graph data, the binary codes for each node can be generated by learning a hash function with the sign function~\cite{tan2020learning} or the binary regularization loss~\cite{tan2020learning} to binarize the hidden representation.
Since our problem is focusing on compressing the \textit{input} embedding rather than the intermediate hidden representation, learning-to-hash methods like the ones proposed by~\cite{tan2020learning,qin2020ghashing} are not applicable in our scenario.
Other compression methods for graph data like the method proposed by~\cite{chen2020differentiable} 
requires the embedding table (i.e., query matrix
) as an input for the training process, and is thus also impractical to perform training on large graphs.
To the best of our knowledge, our work is the first to focus on studying the compression of input embedding for graph data.

\section{Method}
\begin{figure*}[ht]
\begin{center}
\includegraphics[width=0.7\linewidth]{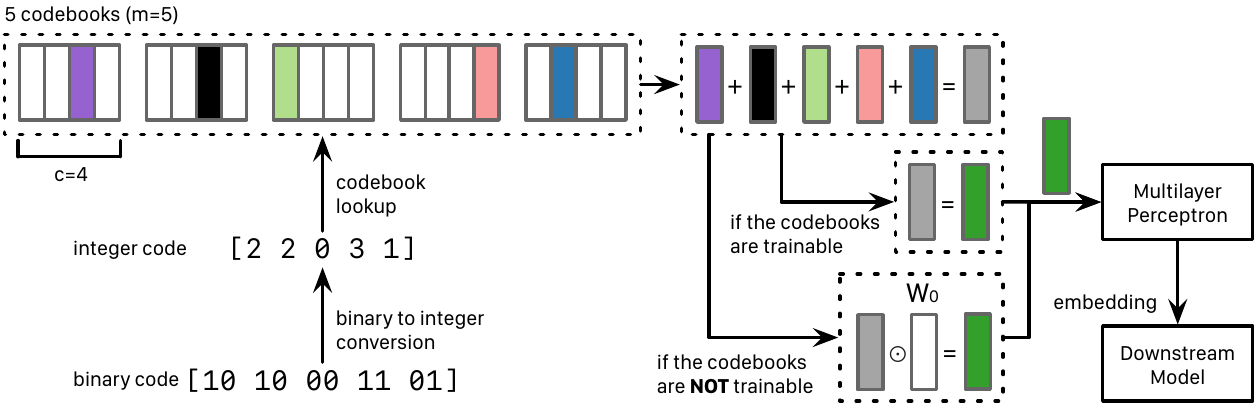}
\end{center}
\caption{
In this toy example, each codebook has 4 vectors ($c=4$) and there are 5 \textit{distinct} codebooks ($m=5$).
There are two variants of the adopted decoder models: 1) a light version where the codebooks are \textbf{NOT} trainable and 2) a full version where the codebooks are trainable.
$W_0$ is a trainable vector for rescaling the intermediate representations (see Section~\ref{sec:decoder}).
}
\label{fig:decoder_design}
\end{figure*}

The proposed method consists of two stages: 1) an encoding stage where each node's compositional code is generated with a hashing-based method and 2) a decoding stage where the decoder is trained in an end-to-end fashion together with the downstream model.
Figure~\ref{fig:decoder_design} shows an example forward pass.
The binary code is a node's compositional code generated by the hashing-based method (Section~\ref{sec:hashing}).
After the binary code is converted to integer code, the decoder model, which mostly includes $m$ codebooks and a multilayer perceptron (MLP) as described in Section~\ref{sec:decoder}, generates the corresponding embedding.
The memory cost of storing both the compositional codes and the decoder is drastically lower than the conventional embedding layer.

\subsection{Hashing-based Coding Scheme}
\label{sec:hashing}
Algorithm~\ref{alg:encode} outlines the random projection-based hashing method.
The first input to our algorithm is a matrix~$\mathbf{A} \in \mathbb{R}^{n \times d}$ containing the auxiliary information of each node where $n$ is the number of nodes and $d$ is the length of auxiliary vector associated with each node.
When the adjacency matrix is used as the auxiliary information, $d$ is equal to $n$, and it is preferred to store~$\mathbf{A}$ as a sparse matrix in compressed row storage (CRS) format as all the operations on $\mathbf{A}$ are row-wise operations.
The other inputs include code cardinality~$c$ and code length~$m$.
These two inputs dictate the format (and the memory cost) of the output compositional code~$\mathbf{\hat{X}}$.
For each node's associated code vector, $c$ controls the cardinality of each element in the code vector and $m$ controls the length of the code vector.
The output is the resulting compositional codes~$\mathbf{\hat{X}} \in \mathbb{B}^{n \times m \log_2 c}$ in the binary format where each row contains a node's associated code vector.
$m \log_2 c$ is the number of bits required to store one code vector and $c$ is the power of 2.
We store~$\mathbf{\hat{X}}$ in binary format because the binary format is more space-efficient compared to the integer format.
The binary code vector can be reversed back to integer format before inputting it to the decoder.

\begin{algorithm}[ht]
    \centering
    \caption{Encode with Random Projection\label{alg:encode}}
    {\footnotesize
    \begin{algorithmic}[1]
        \Input{auxiliary information~$\mathbf{A} \in \mathbb{R}^{n \times d}$, code cardinality~$c$, code length~$m$}
        \Output{compositional code~$\mathbf{\hat{X}} \in \mathbb{B}^{n \times m \log_2 c}$}
        \Function{Encode}{$A, c, m$}
        \State $n_{\text{bit}} \gets m \log_2 c$
        \State $\mathbf{\hat{X}} \gets \textsc{ GetAllFalseBooleanMatrix}(n, n_{\text{bit}})$
        \For{$i$ \textbf{in} $[0, n_{\text{bit}})$}
        \State $V \gets \textsc{ GetRandomVector}(d)$
        \State $U \gets \textsc{ GetEmptyVector}(n)$
        \For{$j$ \textbf{in} $[0, n)$}
        \State $U[j] \gets \textsc{ DotProduct}(\mathbf{A}[j, :], V)$
        \EndFor
        \State $t \gets \textsc{ GetMedian}(U)$
        \For{$j$ \textbf{in} $[0, n)$}
        \State \textbf{if} $U[j] > t$ \textbf{then} $\mathbf{\hat{X}}[j, i] \gets \mathtt{True}$
        \EndFor
        \EndFor
        \State \Return $\mathbf{\hat{X}}$
        \EndFunction
    \end{algorithmic}}
\end{algorithm}

In line 2, the number of bits required to store each code vector (i.e., $m \log_2 c$) is computed and stored in variable~$n_{\text{bit}}$.
In line 3, a Boolean matrix~$\mathbf{\hat{X}}$ of size $n \times n_{\text{bit}}$ is initialized for storing the resulting compositional codes.
The default value for~$\mathbf{\hat{X}}$ is $\mathtt{False}$.
From line 4 to 11, the compositional codes are generated bit-by-bit in the outer loop and node-by-node in the inner loops.
Generating compositional codes in such order is a more memory-efficient way to perform random projection as it only needs to keep a size~$d$ random vector in each iteration compared to the alternative order.
If the inner loop (i.e., line 7 to 8) is switched with the outer loop (i.e., line 4 to 11), it would require us to use a $\mathbb{R}^{n_{\text{bit}} \times d}$ matrix to store all the random vectors for the random projection (i.e., matrix multiplication).

In line 5, a random vector~$V \in \mathbb{R}^d$ is generated; the vector~$V$ is used for performing random projection.
In line 6, a vector~$U \in \mathbb{R}^n$ is initialized for storing the result of random projection.
From line 7 to 8, each node's associated auxiliary vector is projected using the random vector~$V$ and stored in $U$ (i.e., $U = \mathbf{A}V$).
Here, the memory footprint could be further reduced if we only load a few rows of $\mathbf{A}$ during the loop instead of the entire $\mathbf{A}$ before the loop.
Such optimization could be important as the size of $\mathbf{A}$ could be too large for systems with limited memory.
In line 9, the median of $U$ is identified and stored in $t$.
This is the threshold for binarizing real values in $U$.
From line 10 to 11, using both the values in vector~$U$ and the threshold~$t$, the binary code is generated for each node.
Lastly, in line 12, the resulting compositional codes~$\mathbf{\hat{X}}$ are returned.
The resulting~$\mathbf{\hat{X}}$ can be used for any downstream tasks.

\begin{figure}[ht]
\begin{center}
\includegraphics[width=0.9\linewidth]{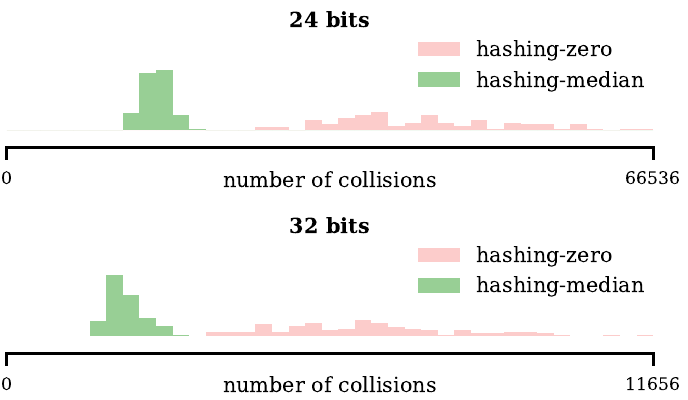}
\end{center}
\caption{
The experiments are performed on \texttt{metapath2vec} for 100 times under two different bit length settings: 24 bits and 32 bits.
The distribution of the 100 outcomes (i.e., number of collisions) for each method is shown in the figure.
The number of collisions is lower for the median threshold compared to the zero threshold.
}
\label{fig:meta_collision_m2v}
\end{figure}

Note, we use the median as the threshold instead of the more commonly seen zero because it reduces the number of collisions in the resulting binary code\footnotemark.
\footnotetext{The threshold used in the LSH method proposed by~\cite{charikar2002similarity} is zero.}
Reducing the number of collisions is important for our case because our goal is to generate a unique code vector to represent each node.
To confirm whether using the median as the threshold reduces the number of collisions, we have performed an experiment using pre-trained \texttt{metapath2vec} node embeddings~\cite{dong2017metapath2vec} as the auxiliary matrix~$\mathbf{A}$.
Pre-trained embeddings provide us a way to quickly benchmark different LSH thresholds.
We do not use pre-trained embeddings in the intended use cases (i.e., Section~\ref{sec:gnn_embedding} and Section~\ref{sec:mci_embedding}).
We generate the compositional codes with random projection-based hashing with either the median or zero as the threshold.
Then, we count the number of collisions in the generated compositional codes.
We repeat the experiment 100 times under two different experimental settings (i.e., 24 bits/32 bits).
The experiment results are summarized in Figure~\ref{fig:meta_collision_m2v} with histograms, setting the threshold to median instead of zero indeed reduces the number of collisions.
We also repeat the experiments with \texttt{metapath2vec++} and \texttt{GloVe} embeddings and the conclusion remains the same.
Please see Appendix~\ref{app:collision} for experiment setup details and additional experiment results.

\sloppy
The memory complexity of Algorithm~\ref{alg:encode} is $O(\textsc{max}(n m \log_2 c, d f, n f))$ where $f$ is the number of bits required to store a floating-point number. 
The $n m \log_2 c$ term is the memory cost associated with storing~$\mathbf{\hat{X}}$, the $d f$ term is the memory cost associated with storing~$V$, and the $n f$ term is the memory cost associated with storing~$U$.
Because $f$ is usually less than $m \log_2 c$ (i.e., based on hyper-parameters used in~\cite{shu2017compressing,takase2020all}\footnotemark) and $d$ is usually less than or equal to $n$, the typical memory complexity of Algorithm~\ref{alg:encode} is $O(n m \log_2 c)$.
\footnotetext{If single-precision format is used for floating-point numbers, $f$ is 32 bit, and $m \log_2 c$ is commonly set to a number larger than 32 bit in~\cite{shu2017compressing,takase2020all}.}
In other words, the memory complexity of Algorithm~\ref{alg:encode} is the same as the output matrix~$\mathbf{\hat{X}}$ which shows how memory efficient Algorithm~\ref{alg:encode} is.
The time complexity of Algorithm~\ref{alg:encode} is $O(n m \log_2 c d)$ for the nested loop\footnotemark.
\footnotetext{The median finding algorithm~\cite{blum1973time} in line 9 is $O(n)$ which is the same as the inner loops (i.e., line 7 to 8 and line 10 to 12.).}

\subsection{Decoder Model Design}
\label{sec:decoder}
We will use the example forward pass presented in Figure~\ref{fig:decoder_design} to introduce the decoder design.
The input to the decoder is the binary compositional codes generated from the hashing-based coding scheme introduced in Section~\ref{sec:hashing}.
The input binary code is first converted to integers for use as indexes for retrieving the corresponding real vector from the codebooks.
In our example, the binary vector $\mathtt{\left[10, 10, 00, 11, 01\right]}$ is converted to integer vector $\mathtt{\left[2, 2, 0, 3, 1\right]}$.
Each codebook is a $\mathbb{R}^{c \times d_c}$ matrix where $c$ is the number of codes in each codebook (i.e., code cardinality) and $d_c$ is the length of each real vector in the codebook.
There are $m$ codebooks in total where $m$ is the code length (i.e., length of the code after being converted to integer vector from binary vector).
Because the code length is 4 in the example, there are 5 codebooks in Figure~\ref{fig:decoder_design}.
Because the code cardinality is 4 (i.e., the number of possible values in the integer code), each codebook has 4 real vectors.

From each codebook, a real number vector is retrieved based on each codebook's corresponding index.
In our example, the vector corresponding to index 2 (purple) is retrieved from the first codebook, index 2 (black) is retrieved from the second codebook, index 0 vector (green) is retrieved from the third codebook, index 3 vector (red) is retrieved from the forth codebook, and the index 1 vector (blue) is retrieved from the last codebook.
The real vectors (i.e., the codebooks) can either be non-trainable random vectors or trainable vectors.
We refer to the former method as the \textit{light} method and the later method as the \textit{full} method.
The former method is lighter as the later method increases the number of trainable parameters by $m c d_c$.
The full method is desired if the additional trainable parameters (i.e., memory cost) are allowed by the hardware.
Note, despite the full method having a higher memory cost, the number of trainable parameters is still independent of the number of nodes in the graph.

Next, the retrieved real vectors are summed together.
The summed vector is handled differently for the light and full methods.
As the codebooks are not trainable for the light method, we compute the element-wise product between the summed vector and a trainable vector~$W_0 \in \mathbb{R}^{d_c}$ to rescale each dimension of the summed vector following~\cite{takase2020all}.
Such transformation is not needed for the full method because it can capture this kind of transformation with the trainable parameters in the codebooks.  
The transformed vector is then fed to an MLP with ReLU between linear layers. 
The output of the MLP is the embedding corresponding to the input compositional code for the downstream model.

If the number of neurons for the MLP is set to $d_m$, the number of layers for the MLP is set to $l$, and the dimension of the output embedding is set to $d_e$, the light method has $m c d_c$ non-trainable parameters (which can be stored outside of GPU memory) and $d_c + d_c d_m + (l - 2) d_m^2 + d_m d_e$ trainable parameters.
The full method has $m c d_c + d_c d_m + (l - 2) d_m^2 + d_m d_e$ trainable parameters.
Here, we assume $l$ is greater than or equal to 2.
Note, the number of parameters does not grow with the increasing number of nodes for both the light and full methods.
The decoder model used in~\cite{takase2020all} is the light method without the binary to integer conversion step.

\section{Integration with GNN Models}
\begin{figure}[ht]
\begin{center}
\includegraphics[width=0.7\columnwidth]{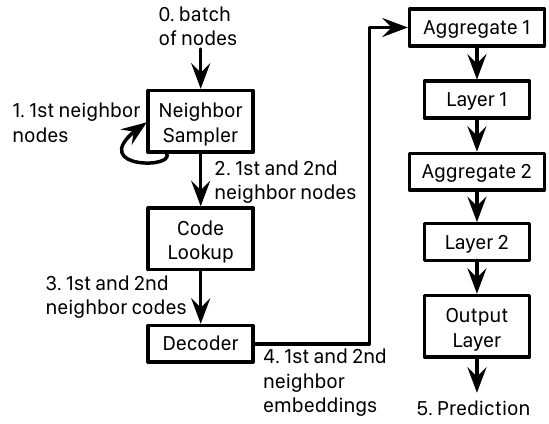}
\end{center}
\caption{
The proposed method can be integrated with the \texttt{GraphSage} model.
The \textit{Code Lookup} is used to look up the corresponding binary code for each input node. 
The \textit{Decoder} is the system presented in Figure~\ref{fig:decoder_design} and converts the input binary codes to embeddings.
}
\label{fig:graphsage}
\end{figure}

In this section, we show how the proposed method can be integrated with the 
GNN models.
Figure ~\ref{fig:graphsage} depicts an example where we use the proposed method with the \texttt{GraphSAGE} model~\cite{hamilton2017inductive}, one of the most prevalent GNN models applied to large scale (e.g. industrial level) data. 
Other GNNs can be integrated with the proposed method in a similar fashion (i.e., by replacing the embedding layer with the proposed method).
First, in step 0, a batch of nodes is sampled.
In step 1, for each node in the batch, a number of neighboring nodes (i.e., first neighbors) are sampled.
Because the example model shown in the figure has 2 layers, the neighbors of neighbors (i.e., second neighbors) are also sampled in step 2.
Next, the binary codes associated with each node's first and second neighbors are retrieved in step 3 and decoded in step 4 using the system described in Section~\ref{sec:decoder}.

After the embeddings for both the first and second neighbors are retrieved, the second neighbor embeddings of each given first neighbor embedding are aggregated with functions like \texttt{mean} or \texttt{max} in \texttt{Aggregate 1} layer.
Let's say $H_i$ contains the embeddings of neighboring nodes for a given node $i$, the aggregate layer computes the output $\hat{h}_i$ with $\textsc{Aggregate}(H_i)$.
Next, in \texttt{Layer 1}, for each first neighbor node $i$, $\hat{h}_i$ and $x_i$ (i.e., embedding for node $i$) are concatenated and processed with a linear layer plus non-linearity.
The process of \texttt{Layer 1} can be represented with  $\sigma(W \cdot \textsc{Concatenate}(\hat{h}_i, x_i))$ where $W$ is the weight associated with the linear layer and $\sigma(\cdot)$ is the non-linear function like \texttt{ReLU}.
A similar process is repeated in \texttt{Aggregate 2} layer and \texttt{Layer 2} to generate the final representation of each node in the batch.
The final prediction is computed by feeding the learned representation to the output (i.e., linear) layer.
All parameters in the model are learned end-to-end using the training data.

\section{Experiment}
\label{sec:experiment}
We perform three sets of experiments: 1) pre-trained embedding reconstruction, 2) training decoder with GNN models jointly for node classification and link prediction problems, and 3) an industrial application of the proposed method with merchant category identification problem~\cite{yeh2020merchant}.
The first set of experiments uses proxy tasks to reveal the difference between different methods' compressing capability while the second set of experiments provides the performance measurement of different methods on common graph problems like node classification and link prediction.
The third set of experiments compares the proposed method with the baseline on an industrial problem.
Experiments are conducted in \texttt{Python} (see~\cite{sourceurl}).

\subsection{Pre-trained Embedding Reconstruction}
\label{sec:pretrain}
In this set of experiments, we compare the compression capability of different compression methods by testing the quality of the reconstructed embedding.
Note, this set of experiments uses proxy tasks as additional test-beds to highlight the difference between different compression methods.
In the intended use scenarios (see Section~\ref{sec:gnn_embedding} and Section~\ref{sec:mci_embedding}), the pre-trained embeddings do not come with the dataset.
The tested methods are the random coding (i.e., baseline method proposed by~\cite{takase2020all}), the learning-based coding (i.e., autoencoder similar to the method proposed by~\cite{shu2017compressing}), and the hashing-based coding (i.e., the proposed method).
When applying the hashing-based coding method on the graph dataset, we feed either the original pre-trained embedding (i.e., hashing/pre-trained in Figure~\ref{fig:pretrain}) or the adjacency matrix from the graph (i.e., hashing/graph in Figure~\ref{fig:pretrain}) into Algorithm~\ref{alg:encode}.
We vary the number of compressed entities when testing different methods.

\subsubsection{Dataset}
\sloppy
Three sets of pre-trained embeddings are used in these experiments: 1) the 300 dimension \texttt{GloVe} word embeddings, 2) the 128 dimension \texttt{metapath2vec} node embeddings, and 3) the 128 dimension \texttt{metapath2vec++} node embeddings.
The \texttt{GloVe} embeddings are tested with word analogy and similarity tasks.
The performance measurements for word analogy and similarity tasks are Accuracy and Spearman's Rank Correlation ($\rho$), respectively.
The \texttt{metapath2vec}/\texttt{metapath2vec++} embeddings are tested with node clustering, and the performance measurement is Normalized Mutual Information.
Please see Appendix~\ref{app:pretrain_dataset} for more details regarding the datasets.

\subsubsection{Implementation}
We use the full decoding method in this set of experiments.
To train the compression method, we use mean squared error between the input embeddings and the reconstructed embeddings as the loss function following~\cite{takase2020all}.
The loss function is optimized with \texttt{AdamW}~\cite{loshchilov2017decoupled} with the default
hyper-parameter settings in \texttt{PyTorch}~\cite{paszke2019pytorch}.
Because we want to vary the numbers of compressed entities when comparing different methods, we need to sample from the available pre-trained embeddings. 
Similar to~\cite{takase2020all}, we sample based on the frequency\footnotemark.
\footnotetext{For \texttt{GloVe}, frequency means the times a word occurs in the training data.
For \texttt{metapath2vec} and \texttt{metapath2vec++}, frequency means the times of a node occurs in the sampled metapaths.}
Since different experiments use different numbers of compressed entities, we only evaluate with the same top 5k entities based on frequency similar to~\cite{takase2020all}, despite there being more than 5k reconstructed embeddings when the number of compressed entities is greater than 5k.
In this way, we have the same test data across experiments with different numbers of compressed entities.
The detailed hyper-parameter settings are shown in Appendix~\ref{app:pretrain_hyper}.

\subsubsection{Result}
The experiment results are summarized in Figure~\ref{fig:pretrain}.
Note, we use ``random" to denote the baseline method (i.e., \texttt{ALONE}).
When the number of compressed entities is low, the reconstructed embeddings from all compression methods perform similar to using the raw embeddings (i.e., the original pre-trained embeddings).
As the number of compressed entities increases, the reconstructed embeddings' performance decreases.
The decreasing performance is likely caused by the fact that the decoder model's size does not grow with the number of compressed entities.
In other words, the compression ratio increases as the number of compressed entities increases (see Table~\ref{tab:compression}).
When comparing different compression methods, we can observe that the quality of the reconstructed embeddings from the random coding method drops sharply compared to other methods (i.e., hashing-based coding and learning-based coding).
It is surprising that the hashing-based coding method works as well as the learning-based coding method even if the learning-based coding method uses additional parameters to learn the coding function.
When we compare both variants of the proposed coding method (i.e., hashing with pre-trained and hashing with graph/adjacency matrix), the performance is very similar.
This shows how the adjacency matrix from the graph is a valid choice for applying the proposed hashing-based coding method.
We have also tested other settings of $c$ and $m$ (see Table~\ref{tab:pretrain_add} and accompany text); the conclusion stays the same.
Overall, the hashing-based coding method outperforms the baseline \texttt{ALONE} method.

\subsection{Node Classification and Link Prediction}
\label{sec:gnn_embedding}
To examine the difference between the compression methods on graph related tasks, we perform node classification and link prediction experiments where the decoder is trained together with GNNs, e.g.  \texttt{GraphSAGE}~\cite{hamilton2017inductive}, Graph Convolutional Network (i.e., \texttt{GCN})~\cite{kipf2016semi}, Simplifying Graph Convolutional Network (i.e., \texttt{SGC})~\cite{wu2019simplifying}, and Graph Isomorphism Network (i.e., \texttt{GIN})~\cite{xu2018powerful}.
Because we assume there is no node features or pre-trained embeddings available in our experiment setup, the autoencoder-based method proposed by~\cite{shu2017compressing} is not applicable.
We compare the proposed hashing-based coding method (using adjacency matrices and Algorithm~\ref{alg:encode} to generate the code) with two baseline methods: random coding method and raw embedding method.
The raw embedding method explicitly learns the embeddings together with the GNN model, which can be treated as the upper bound in terms of accuracy because the embeddings are not compressed.

\subsubsection{Dataset}
The experiments are performed on the ogbn-arxiv, ogbn-mag, ogbn-products, ogbl-collab, and ogbl-ddi datasets from Open Graph Benchmark~\cite{hu2020open}.
As we are more interested in evaluating our model performance in attribute-less graphs, we use only graph structure information of these datasets.
We convert all the directed graphs to undirected graphs by making the adjacency matrix symmetry.
The ogbn-mag dataset is a heterogeneous graph, and we only use the citing relation between paper nodes as the labels are associated with paper nodes.
The performance measurement is accuracy for node classification datasets, hits@50 for ogbl-collab, and hits@20 for ogbl-ddi.

\subsubsection{Implementation}
We use the \texttt{PyTorch} implementation of the \texttt{GraphSAGE} model with mean pooling aggregator~\cite{hamilton2017inductive,pytorchgraphsage}.
We use PyG library~\cite{fey2019fast} to implement \texttt{GCN}~\cite{kipf2016semi}, \texttt{SGC}~\cite{wu2019simplifying}, and \texttt{GIN}~\cite{xu2018powerful}.
The model parameters are optimized by \texttt{AdamW}~\cite{loshchilov2017decoupled} with the default hyper-parameter settings.
The detailed hyper-parameter settings are shown in Appendix~\ref{app:gnn_hyper}.

\subsubsection{Result}
The experimental results are shown in Table~\ref{tab:graph_acc_gcn2}. 
The results show that the hashing-based coding method outperforms the random coding method (i.e., \texttt{ALONE}) in most tested scenarios.
One possible reason for the random coding method's less impressive performance compared to the result reported on NLP tasks by~\cite{takase2020all} is related to the number of entities compressed by the compression method.
In NLP models, embeddings typically represent sub-words instead of words~\cite{vaswani2017attention}.
For example, the transformer model for machine translation adopted by~\cite{takase2020all} has 32,000 sub-words, which is much smaller compared to most tested graph datasets (i.e., ogbn-arxiv, ogbn-mag, ogbn-products, and ogbl-collab all have more than 150,000 nodes).
In other words, the proposed hashing-based coding method is more effective for representing a larger set of entities in compressed space compared to the baseline random coding method.
The proposed method is also compared to the ``without compression" baseline (i.e., NC).
The NC baseline outperforms the proposed method in 10 out of 20 experiments as expected since the compression used in our method is lossy.
One possible reason for the ten unexpected outcomes (i.e., the proposed method outperforms the NC baseline) is that the lossy compression may sometimes remove the ``correct" noise from the data.
Because the focus of this paper is scalability when comparing to the NC baseline, we left the study of such phenomena for future work.

\begin{table}[htb]
\caption{
The proposed hashing-based coding almost always outperforms the baseline random coding with different GNNs for both node classification and link prediction. It also achieves close to, and occasionally outperforms, the non-compressed method.
We use NC to denote the non-compressed or embedding learning method \textit{without} compression, Rand to denote the random coding method (i.e., \texttt{ALONE}), and Hash to denote the proposed hashing coding method.
}
\begin{center}
\footnotesize

\begin{tabular}{ll||ccc}
\multirow{2}{*}{task} & \multirow{2}{*}{dataset} & \multicolumn{3}{c}{\texttt{GraphSage}} \\
& & \multicolumn{1}{c|}{NC} & Rand & Hash \\ \hline \hline
\multirow{3}{*}{node classification} & ogbn-arxiv (acc.) & \multicolumn{1}{c|}{0.6228} & 0.6045 & \textbf{0.6259} \\
& ogbn-mag (acc.) & \multicolumn{1}{c|}{0.3192} & 0.2989 & \textbf{0.3387} \\
& ogbn-products (acc.) & \multicolumn{1}{c|}{0.7486} & 0.6327 & \textbf{0.6414} \\ \hline 
\multirow{2}{*}{link prediction} & ogbl-collab (hits@50) & \multicolumn{1}{c|}{0.2740} & \textbf{0.1966} & 0.1956 \\
 & ogbl-ddi (hits@20) & \multicolumn{1}{c|}{0.3277} & 0.3043 & \textbf{0.3429}
\end{tabular}

\vspace{0.1cm}

\begin{tabular}{ll||ccc}
\multirow{2}{*}{task} & \multirow{2}{*}{dataset} & \multicolumn{3}{c}{\texttt{GCN}} \\
& & \multicolumn{1}{c|}{NC} & Rand & Hash \\ \hline \hline
\multirow{3}{*}{node classification} & ogbn-arxiv (acc.) & \multicolumn{1}{c|}{0.5251} & 0.4957 & \textbf{0.5437} \\
& ogbn-mag (acc.) & \multicolumn{1}{c|}{0.1815} & 0.1146 & \textbf{0.3466} \\
& ogbn-products (acc.) & \multicolumn{1}{c|}{0.4719} & 0.3594 & \textbf{0.4914} \\ \hline 
\multirow{2}{*}{link prediction} & ogbl-collab (hits@50) & \multicolumn{1}{c|}{0.2316} & 0.1647 & \textbf{0.1898} \\
& ogbl-ddi (hits@20) & \multicolumn{1}{c|}{0.3697} & \textbf{0.3399} & 0.3319
\end{tabular}

\vspace{0.1cm}

\begin{tabular}{ll||ccc}
\multirow{2}{*}{task} & \multirow{2}{*}{dataset} & \multicolumn{3}{c}{\texttt{SGC}} \\
& & \multicolumn{1}{c|}{NC} & Rand & Hash \\ \hline \hline
\multirow{3}{*}{node classification} & ogbn-arxiv (acc.) & \multicolumn{1}{c|}{0.6690} & 0.5491 & \textbf{0.5809} \\
& ogbn-mag (acc.) & \multicolumn{1}{c|}{0.3523} & 0.1839 & \textbf{0.3657} \\
& ogbn-products (acc.) & \multicolumn{1}{c|}{0.7686} & 0.3767 & \textbf{0.4966} \\ \hline 
\multirow{2}{*}{link prediction} & ogbl-collab (hits@50) & \multicolumn{1}{c|}{0.5589} & 0.4790 & \textbf{0.5116} \\
& ogbl-ddi (hits@20) & \multicolumn{1}{c|}{0.4841} & 0.5575 & \textbf{0.5941}
\end{tabular}

\vspace{0.1cm}

\begin{tabular}{ll||ccc}
\multirow{2}{*}{task} & \multirow{2}{*}{dataset} & \multicolumn{3}{c}{\texttt{GIN}} \\
& & \multicolumn{1}{c|}{NC} & Rand & Hash \\ \hline \hline
\multirow{3}{*}{node classification} & ogbn-arxiv (acc.) & \multicolumn{1}{c|}{0.5546} & 0.3736 & \textbf{0.5263} \\
& ogbn-mag (acc.) & \multicolumn{1}{c|}{0.2728} & 0.2011 & \textbf{0.3414} \\
& ogbn-products (acc.) & \multicolumn{1}{c|}{0.6423} & 0.4396 & \textbf{0.5706} \\ \hline 
\multirow{2}{*}{link prediction} & ogbl-collab (hits@50) & \multicolumn{1}{c|}{0.2614} & 0.2086 & \textbf{0.2475} \\
& ogbl-ddi (hits@20) & \multicolumn{1}{c|}{0.3216} & 0.3536 & \textbf{0.3876}
\end{tabular}

\label{tab:graph_acc_gcn2}
\end{center}
\end{table}

In terms of memory usage, the compression method is capable of achieving a considerably good compression ratio.
For example, since the ogbn-products dataset has 1,871,031 nodes, it requires 456.79 MB to store the raw embeddings in GPU memory.
On the contrary, it only takes the proposed method 28.55 MB to store the binary codes in CPU memory, and the corresponding decoder model only costs 9.13 MB of GPU memory.  
The compression ratio is 43.75 for the proposed method's less memory efficient setup (i.e., full model) if we only consider GPU memory usage.
For the total memory usage, the compression ratio is 11.74 for the same setup.
The complete memory cost breakdown is shown in Table~\ref{tab:graph_products}.
The unit for memory is megabytes (MB), and the column label ``ratio" stands for ``compression ratio". 

\begin{table*}[tb]
\caption{The memory cost (MB) for models on ogbn-products dataset.}
\begin{center}
\footnotesize
\begin{tabular}{l||cc|c||cc|cc||cc}
\multirow{2}{*}{Method} & \multicolumn{3}{c||}{CPU} & \multicolumn{4}{c||}{GPU} & \multicolumn{2}{c}{CPU+GPU} \\
 & \begin{tabular}[c]{@{}c@{}}Binary\\      Code\end{tabular} & Decoder & Total & \begin{tabular}[c]{@{}c@{}}Decoder or\\      Embedding\end{tabular} & GNN & Total & Ratio & Total & Ratio \\ \hline \hline
Raw & 0.00 & 0.00 & 0.00 & 456.79 & 1.35 & 458.14 & 1.00 & 458.14 & 1.00 \\
Hash-Light & 28.55 & 8.00 & 36.55 & 1.13 & 1.35 & 2.47 & 185.34 & 39.02 & 11.74 \\
Hash-Heavy & 28.55 & 0.00 & 28.55 & 9.13 & 1.35 & 10.47 & 43.75 & 39.02 & 11.74
\end{tabular}
\label{tab:graph_products}
\end{center}
\end{table*}

\subsection{Merchant Category Identification}
\label{sec:mci_embedding}
In this section, we use a real-world application to evaluate the effectiveness of the proposed embedding compression method compared to the baseline. 
In this case, we apply our model to a large scale transaction graph.
The transaction volume of credit and debit card payments has proliferated in recent years with the rapid growth of small businesses and online retailers. 
When processing these payment transactions, recognizing each merchant's real identity (i.e., merchant category or business type) is vital to ensure the integrity of payment processing systems as a merchant could falsify its identity by registering in an incorrect merchant category within payment processing companies.
For example, a high-risk merchant may pretend to be in a low-risk merchant category by reporting a fake merchant category to avoid higher processing fees associated with risky categories.
Specific business type (i.e., gambling) is only permitted in some regions and territories.
A merchant could report a false merchant category to avoid scrutiny from banks and regulators.
A merchant may also report the wrong merchant category by mistake.

\begin{figure}[ht]
\begin{center}
\includegraphics[width=0.85\linewidth]{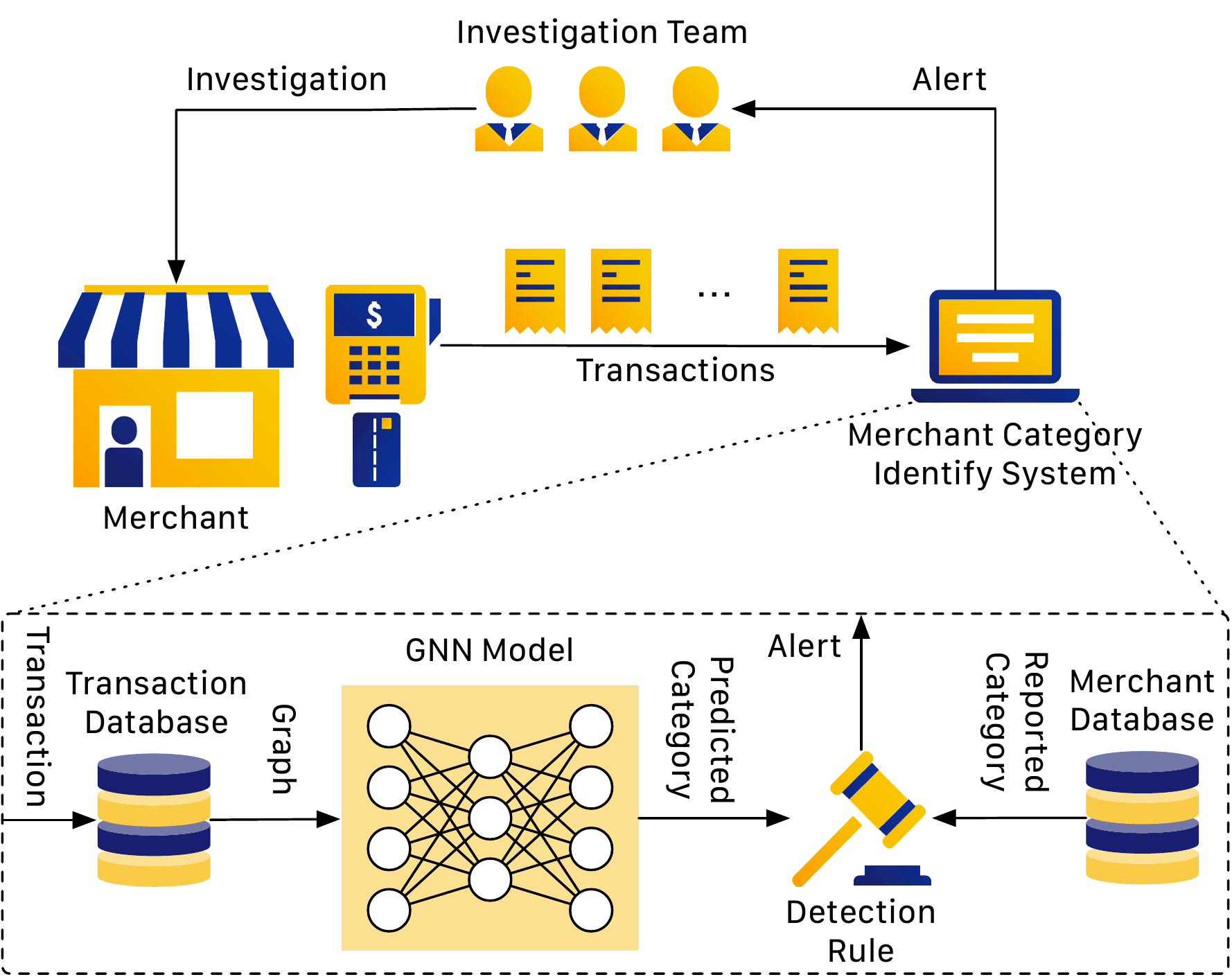}\\
\end{center}
\caption{
The overall design for the merchant category identification system.
The proposed method is used in the GNN model component in the system.
}
\label{fig:mcc_id_sys}
\end{figure}

We use the system depicted in Figure~\ref{fig:mcc_id_sys} to identify merchants with possible faulty merchant categories.
The merchant category identification system monitors the transactions of each merchant and notifies the investigation team whenever the identified merchant category mismatches with the merchant’s self-reported category.
We represent the transaction data in a consumer-merchant bipartite graph and use the GNN-based classification model to identify a merchant's category. 
The performance of the classification model dictates the success of the overall identification system.
As there are millions of consumers and merchants using the payment network, we want to build a scalable and accurate GNN model for the identification system.
To achieve this goal, we compare the proposed hashing-based coding scheme with the baseline random coding scheme using real transaction data.

\subsubsection{Dataset}
We create a graph dataset by sampling transactions from January to August in 2020.
The resulting graph dataset consists of 17,943,972 nodes; of which 9,089,039 are consumer nodes, and 8,854,933 are merchant nodes.
There is a total of 651 merchant categories in the dataset.
We use 70\% of the merchant nodes as the training data, 10\% of the merchant nodes as the validation data, and 20\% of the merchant nodes as the test data.
Because the classification model is solving a multi-class classification problem, we use accuracy (i.e., acc.) as the performance measurement.
We also report the hit rate at different thresholds (i.e., hit@$k$) as the particular detection rule implemented in the identification system has its performance tie strongly with the hit rate of the model.

\subsubsection{Implementation}
We use the following hyper-parameter settings for the decoders: $l=3$, $d_c=d_m=512$, $d_e=64$, $c=256$, and $m=16$.
Because of the sheer size of the dataset, it is impossible to run the non-compressed baseline on this dataset.
We once again use the \texttt{PyTorch} implementation of the \texttt{GraphSAGE} model with mean pooling aggregator~\cite{hamilton2017inductive,pytorchgraphsage}.
We choose to move forward with the \texttt{GraphSAGE} model because it provides the best node classification performance from previous experiments.
We use the following hyper-parameter settings for the \texttt{GraphSAGE} model: $\text{number of layers}=2$, $\text{number of neurons}=128$, $\text{activation function}=\texttt{ReLU}$, and $\text{number of neighbors}=5$.
We use the following hyper-parameter settings for the \texttt{AdamW} optimizer~\cite{loshchilov2017decoupled} when optimizing the cross entropy loss function: $\text{learning rate}=0.01$, $\beta_1=0.9$, $\beta_2=0.999$, and $\text{weight decay}=0$.
We train \texttt{GraphSAGE} for 20 epochs with a batch size of 1024 and report the evaluation accuracy from the epoch with the best validation accuracy.

\subsubsection{Result}
Table~\ref{tab:mcc_id_result} summarizes the experiment result.
As expected, we observe performance gains in all performance measurements by using the proposed hashing-based coding method instead of the random coding method, similar to the experiments presented in prior sections.
The proposed model achieved over 10\% improvement in terms of accuracy when applied in the merchant category identification system.
In addition, with the help of the embedding compression method, it only takes around 2.14 GB to store both binary codes and the decoder.
The performance improvement of the hashing-based method over the baseline random coding method is mild compared to the result presented in Table~\ref{tab:graph_acc_gcn2}.
One possible reason for such observation is that the merchant category identification problem is more difficult than the tasks presented in Section~\ref{sec:gnn_embedding} due to various types of data imbalance issues.
For example, the restaurant category has over 100k merchants while the ambulance service category has less than 1k merchants. 
There are merchants visited by almost one million consumers, but there are also merchants visited by less than one hundred consumers.
Nevertheless, the improvements in accuracy and hit rate are non-trivial, not to mention the drastic reduction of memory cost.

\begin{table}[ht]
\caption{
Comparison of the classification model using different compression methods.
We use Rand to denote the random coding method (i.e., \texttt{ALONE}), and Hash to denote the proposed hashing-based coding method.
}
\begin{center}
\footnotesize
\begin{tabular}{l||cccc}
Method & acc. & hit@5 & hit@10 & hit@20 \\ \hline \hline
Rand & 0.1239 & 0.3725 & 0.4953 & 0.6233 \\
Hash & \textbf{0.1364} & \textbf{0.3867} & \textbf{0.5098} & \textbf{0.6350} \\ \hline \hline
\% improve & 10.09\% & 3.81\% & 2.93\% & 1.88\%
\end{tabular}
\label{tab:mcc_id_result}
\end{center}
\end{table}

\section{Conclusion}
In this work, we proposed a hashing-based coding scheme that generates compositional codes for compactly representing nodes in graph datasets.
The proposed coding scheme outperforms the prior embedding compressing method which uses a random coding scheme in almost all experiments.
On top of that, the performance degradation coming from the lossy compression is minimal as demonstrated in experiment results.
Because the proposed embedding compressing method drastically reduces the memory cost associated with embedding learning, it is now possible to jointly train unique embeddings for all the nodes with GNN models on industrial scale graph datasets as demonstrated in Section~\ref{sec:experiment}.

\subsection{Potential Impact and Future Directions}
Aside from GNNs, the proposed methods can also be combined with other kinds of models on tasks that require learning embeddings for a large set of entities.
For example, it is common to have categorical features/variables with high cardinalities in financial technology/targeted advertising datasets, and embeddings are often used to represent these categorical features~\cite {du2019pcard,yeh2020towards,deng2021deeplight}.
The proposed method is well suited for building memory-efficient deep learning models with these types of large-scale datasets, e.g., for click-through rate (CTR) prediction or recommendation systems.
As a result, the proposed embedding compressing method could potentially address the scalability problems associated with high-cardinality categorical features in many real-world machine learning problems.
Determining the most effective auxiliary information for generating the binary codes should be an interesting direction to explore for different applications.
For example, one practical adjustment could be to use higher-order adjacency matrices to replace the original adjacency matrix since the higher-order auxiliary information, which captures connectivity information on a broader scope, could result in better embedding compression.

\bibliographystyle{ACM-Reference-Format}
\bibliography{ref_short}


\begin{thebibliography}{43}


\ifx \showCODEN    \undefined \def \showCODEN     #1{\unskip}     \fi
\ifx \showDOI      \undefined \def \showDOI       #1{#1}\fi
\ifx \showISBNx    \undefined \def \showISBNx     #1{\unskip}     \fi
\ifx \showISBNxiii \undefined \def \showISBNxiii  #1{\unskip}     \fi
\ifx \showISSN     \undefined \def \showISSN      #1{\unskip}     \fi
\ifx \showLCCN     \undefined \def \showLCCN      #1{\unskip}     \fi
\ifx \shownote     \undefined \def \shownote      #1{#1}          \fi
\ifx \showarticletitle \undefined \def \showarticletitle #1{#1}   \fi
\ifx \showURL      \undefined \def \showURL       {\relax}        \fi
\providecommand\bibfield[2]{#2}
\providecommand\bibinfo[2]{#2}
\providecommand\natexlab[1]{#1}
\providecommand\showeprint[2][]{arXiv:#2}

\bibitem[Blum et~al\mbox{.}(1973)]%
        {blum1973time}
\bibfield{author}{\bibinfo{person}{Manuel Blum}, \bibinfo{person}{Robert~W.
  Floyd}, \bibinfo{person}{Vaughan~R. Pratt}, \bibinfo{person}{Ronald~L.
  Rivest}, \bibinfo{person}{Robert~Endre Tarjan}, {et~al\mbox{.}}}
  \bibinfo{year}{1973}\natexlab{}.
\newblock \showarticletitle{Time bounds for selection}.
\newblock \bibinfo{journal}{\emph{J. Comput. Syst. Sci.}}
  (\bibinfo{year}{1973}).
\newblock


\bibitem[Carter and Wegman(1979)]%
        {carter1979universal}
\bibfield{author}{\bibinfo{person}{J~Lawrence Carter} {and}
  \bibinfo{person}{Mark~N Wegman}.} \bibinfo{year}{1979}\natexlab{}.
\newblock \showarticletitle{Universal classes of hash functions}.
\newblock \bibinfo{journal}{\emph{Journal of computer and system sciences}}
  \bibinfo{volume}{18}, \bibinfo{number}{2} (\bibinfo{year}{1979}),
  \bibinfo{pages}{143--154}.
\newblock


\bibitem[Charikar(2002)]%
        {charikar2002similarity}
\bibfield{author}{\bibinfo{person}{Moses~S Charikar}.}
  \bibinfo{year}{2002}\natexlab{}.
\newblock \showarticletitle{Similarity estimation techniques from rounding
  algorithms}. In \bibinfo{booktitle}{\emph{STOC}}. \bibinfo{pages}{380--388}.
\newblock


\bibitem[Chen et~al\mbox{.}(2020)]%
        {chen2020differentiable}
\bibfield{author}{\bibinfo{person}{Ting Chen}, \bibinfo{person}{Lala Li}, {and}
  \bibinfo{person}{Yizhou Sun}.} \bibinfo{year}{2020}\natexlab{}.
\newblock \showarticletitle{Differentiable product quantization for end-to-end
  embedding compression}. In \bibinfo{booktitle}{\emph{ICML}}.
\newblock


\bibitem[Conneau et~al\mbox{.}(2017)]%
        {wordsimilarity}
\bibfield{author}{\bibinfo{person}{Alexis Conneau}, \bibinfo{person}{Guillaume
  Lample}, \bibinfo{person}{Marc'Aurelio Ranzato}, \bibinfo{person}{Ludovic
  Denoyer}, {and} \bibinfo{person}{Herv{\'e} J{\'e}gou}.}
  \bibinfo{year}{2017}\natexlab{}.
\newblock \bibinfo{title}{Link for word similarity tasks}.
\newblock
\newblock
\newblock
\shownote{\url{https://dl.fbaipublicfiles.com/arrival/wordsim.tar.gz}}.


\bibitem[Deng et~al\mbox{.}(2021)]%
        {deng2021deeplight}
\bibfield{author}{\bibinfo{person}{Wei Deng}, \bibinfo{person}{Junwei Pan},
  \bibinfo{person}{Tian Zhou}, \bibinfo{person}{Deguang Kong},
  \bibinfo{person}{Aaron Flores}, {and} \bibinfo{person}{Guang Lin}.}
  \bibinfo{year}{2021}\natexlab{}.
\newblock \showarticletitle{DeepLight: Deep Lightweight Feature Interactions
  for Accelerating CTR Predictions in Ad Serving}. In
  \bibinfo{booktitle}{\emph{WSDM}}. \bibinfo{pages}{922--930}.
\newblock


\bibitem[Dong et~al\mbox{.}(2017a)]%
        {metapath2vec}
\bibfield{author}{\bibinfo{person}{Yuxiao Dong}, \bibinfo{person}{Nitesh~V
  Chawla}, {and} \bibinfo{person}{Ananthram Swami}.}
  \bibinfo{year}{2017}\natexlab{a}.
\newblock \bibinfo{title}{Link for {metapath2vec}}.
\newblock
\newblock
\newblock
\shownote{\url{https://ericdongyx.github.io/metapath2vec/m2v.html}}.


\bibitem[Dong et~al\mbox{.}(2017b)]%
        {dong2017metapath2vec}
\bibfield{author}{\bibinfo{person}{Yuxiao Dong}, \bibinfo{person}{Nitesh~V
  Chawla}, {and} \bibinfo{person}{Ananthram Swami}.}
  \bibinfo{year}{2017}\natexlab{b}.
\newblock \showarticletitle{metapath2vec: Scalable representation learning for
  heterogeneous networks}. In \bibinfo{booktitle}{\emph{SIGKDD}}.
\newblock


\bibitem[Du et~al\mbox{.}(2019)]%
        {du2019pcard}
\bibfield{author}{\bibinfo{person}{Min Du}, \bibinfo{person}{Robert
  Christensen}, \bibinfo{person}{Wei Zhang}, {and} \bibinfo{person}{Feifei
  Li}.} \bibinfo{year}{2019}\natexlab{}.
\newblock \showarticletitle{Pcard: personalized restaurants recommendation from
  card payment transaction records}. In \bibinfo{booktitle}{\emph{WWW}}.
\newblock


\bibitem[Duong et~al\mbox{.}(2019)]%
        {duong2019node}
\bibfield{author}{\bibinfo{person}{Chi~Thang Duong}, \bibinfo{person}{Thanh~Dat
  Hoang}, \bibinfo{person}{Ha~The~Hien Dang}, \bibinfo{person}{Quoc Viet~Hung
  Nguyen}, {and} \bibinfo{person}{Karl Aberer}.}
  \bibinfo{year}{2019}\natexlab{}.
\newblock \showarticletitle{On node features for graph neural networks}.
\newblock \bibinfo{journal}{\emph{arXiv preprint arXiv:1911.08795}}
  (\bibinfo{year}{2019}).
\newblock


\bibitem[Faruqui and Dyer(2014)]%
        {faruqui2014community}
\bibfield{author}{\bibinfo{person}{Manaal Faruqui} {and} \bibinfo{person}{Chris
  Dyer}.} \bibinfo{year}{2014}\natexlab{}.
\newblock \showarticletitle{Community evaluation and exchange of word vectors
  at {wordvectors.org}}. In \bibinfo{booktitle}{\emph{ACL: System
  Demonstrations}}. \bibinfo{pages}{19--24}.
\newblock


\bibitem[Fey and Lenssen(2019)]%
        {fey2019fast}
\bibfield{author}{\bibinfo{person}{Matthias Fey} {and}
  \bibinfo{person}{Jan~Eric Lenssen}.} \bibinfo{year}{2019}\natexlab{}.
\newblock \showarticletitle{Fast graph representation learning with {PyTorch
  Geometric}}.
\newblock \bibinfo{journal}{\emph{arXiv preprint arXiv:1903.02428}}
  (\bibinfo{year}{2019}).
\newblock


\bibitem[Hamilton et~al\mbox{.}(2017)]%
        {hamilton2017inductive}
\bibfield{author}{\bibinfo{person}{Will Hamilton}, \bibinfo{person}{Zhitao
  Ying}, {and} \bibinfo{person}{Jure Leskovec}.}
  \bibinfo{year}{2017}\natexlab{}.
\newblock \showarticletitle{Inductive representation learning on large graphs}.
\newblock \bibinfo{journal}{\emph{NeurIPS}}  \bibinfo{volume}{30}
  (\bibinfo{year}{2017}).
\newblock


\bibitem[He et~al\mbox{.}(2020)]%
        {he2020lightgcn}
\bibfield{author}{\bibinfo{person}{Xiangnan He}, \bibinfo{person}{Kuan Deng},
  \bibinfo{person}{Xiang Wang}, \bibinfo{person}{Yan Li},
  \bibinfo{person}{Yongdong Zhang}, {and} \bibinfo{person}{Meng Wang}.}
  \bibinfo{year}{2020}\natexlab{}.
\newblock \showarticletitle{Lightgcn: Simplifying and powering graph
  convolution network for recommendation}. In
  \bibinfo{booktitle}{\emph{SIGIR}}. \bibinfo{pages}{639--648}.
\newblock


\bibitem[He et~al\mbox{.}(2017)]%
        {he2017neural}
\bibfield{author}{\bibinfo{person}{Xiangnan He}, \bibinfo{person}{Lizi Liao},
  \bibinfo{person}{Hanwang Zhang}, \bibinfo{person}{Liqiang Nie},
  \bibinfo{person}{Xia Hu}, {and} \bibinfo{person}{Tat-Seng Chua}.}
  \bibinfo{year}{2017}\natexlab{}.
\newblock \showarticletitle{Neural collaborative filtering}. In
  \bibinfo{booktitle}{\emph{WWW}}.
\newblock


\bibitem[Hu et~al\mbox{.}(2020)]%
        {hu2020open}
\bibfield{author}{\bibinfo{person}{Weihua Hu}, \bibinfo{person}{Matthias Fey},
  \bibinfo{person}{Marinka Zitnik}, \bibinfo{person}{Yuxiao Dong},
  \bibinfo{person}{Hongyu Ren}, \bibinfo{person}{Bowen Liu},
  \bibinfo{person}{Michele Catasta}, {and} \bibinfo{person}{Jure Leskovec}.}
  \bibinfo{year}{2020}\natexlab{}.
\newblock \showarticletitle{Open graph benchmark: Datasets for machine learning
  on graphs}.
\newblock \bibinfo{journal}{\emph{NeurIPS}}  \bibinfo{volume}{33}
  (\bibinfo{year}{2020}), \bibinfo{pages}{22118--22133}.
\newblock


\bibitem[Johnson et~al\mbox{.}(2018)]%
        {pytorchgraphsage}
\bibfield{author}{\bibinfo{person}{Ben Johnson}, \bibinfo{person}{William~L
  Hamilton}, {and} \bibinfo{person}{Can~G\"uney Aksakalli}.}
  \bibinfo{year}{2018}\natexlab{}.
\newblock \bibinfo{title}{pytorch-graphsage}.
\newblock
  \bibinfo{howpublished}{\url{https://github.com/bkj/pytorch-graphsage}}.
\newblock


\bibitem[Kipf and Welling(2017)]%
        {kipf2016semi}
\bibfield{author}{\bibinfo{person}{Thomas~N Kipf} {and} \bibinfo{person}{Max
  Welling}.} \bibinfo{year}{2017}\natexlab{}.
\newblock \showarticletitle{Semi-supervised classification with graph
  convolutional networks}. In \bibinfo{booktitle}{\emph{ICLR}}.
\newblock


\bibitem[Lample et~al\mbox{.}(2018)]%
        {conneau2017word}
\bibfield{author}{\bibinfo{person}{Guillaume Lample}, \bibinfo{person}{Alexis
  Conneau}, \bibinfo{person}{Marc'Aurelio Ranzato}, \bibinfo{person}{Ludovic
  Denoyer}, {and} \bibinfo{person}{Herv{\'e} J{\'e}gou}.}
  \bibinfo{year}{2018}\natexlab{}.
\newblock \showarticletitle{Word translation without parallel data}. In
  \bibinfo{booktitle}{\emph{ICLR}}.
\newblock


\bibitem[{Lample et al.}(2018)]%
        {lample2017unsupervised}
\bibfield{author}{\bibinfo{person}{{Lample et al.}}}
  \bibinfo{year}{2018}\natexlab{}.
\newblock \showarticletitle{Unsupervised Machine Translation Using Monolingual
  Corpora Only}. In \bibinfo{booktitle}{\emph{ICLR}}.
\newblock


\bibitem[Lloyd(1982)]%
        {lloyd1982least}
\bibfield{author}{\bibinfo{person}{Stuart Lloyd}.}
  \bibinfo{year}{1982}\natexlab{}.
\newblock \showarticletitle{Least squares quantization in PCM}.
\newblock \bibinfo{journal}{\emph{IEEE transactions on information theory}}
  \bibinfo{volume}{28}, \bibinfo{number}{2} (\bibinfo{year}{1982}),
  \bibinfo{pages}{129--137}.
\newblock


\bibitem[Loshchilov and Hutter(2018)]%
        {loshchilov2017decoupled}
\bibfield{author}{\bibinfo{person}{Ilya Loshchilov} {and}
  \bibinfo{person}{Frank Hutter}.} \bibinfo{year}{2018}\natexlab{}.
\newblock \showarticletitle{Decoupled Weight Decay Regularization}. In
  \bibinfo{booktitle}{\emph{ICLR}}.
\newblock


\bibitem[Mikolov et~al\mbox{.}(2013)]%
        {word2vec}
\bibfield{author}{\bibinfo{person}{Tomas Mikolov}, \bibinfo{person}{Ilya
  Sutskever}, \bibinfo{person}{Kai Chen}, \bibinfo{person}{Greg~S Corrado},
  {and} \bibinfo{person}{Jeff Dean}.} \bibinfo{year}{2013}\natexlab{}.
\newblock \bibinfo{title}{Link for {word2vec}}.
\newblock
\newblock
\newblock
\shownote{\url{https://code.google.com/archive/p/word2vec/}}.


\bibitem[{Mikolov et al.}(2013)]%
        {mikolov2013distributed}
\bibfield{author}{\bibinfo{person}{{Mikolov et al.}}}
  \bibinfo{year}{2013}\natexlab{}.
\newblock \showarticletitle{Distributed representations of words and phrases
  and their compositionality}. In \bibinfo{booktitle}{\emph{NeurIPS}}.
\newblock


\bibitem[{Paszke et al.}(2019)]%
        {paszke2019pytorch}
\bibfield{author}{\bibinfo{person}{{Paszke et al.}}}
  \bibinfo{year}{2019}\natexlab{}.
\newblock \showarticletitle{Pytorch: An imperative style, high-performance deep
  learning library}.
\newblock \bibinfo{journal}{\emph{NeurIPS}}  \bibinfo{volume}{32}
  (\bibinfo{year}{2019}), \bibinfo{pages}{8026--8037}.
\newblock


\bibitem[Pennington et~al\mbox{.}(2014a)]%
        {pennington2014glove}
\bibfield{author}{\bibinfo{person}{Jeffrey Pennington},
  \bibinfo{person}{Richard Socher}, {and} \bibinfo{person}{Christopher~D
  Manning}.} \bibinfo{year}{2014}\natexlab{a}.
\newblock \showarticletitle{Glove: Global vectors for word representation}. In
  \bibinfo{booktitle}{\emph{EMNLP}}. \bibinfo{pages}{1532--1543}.
\newblock


\bibitem[Pennington et~al\mbox{.}(2014b)]%
        {glove6B}
\bibfield{author}{\bibinfo{person}{Jeffrey Pennington},
  \bibinfo{person}{Richard Socher}, {and} \bibinfo{person}{Christopher~D
  Manning}.} \bibinfo{year}{2014}\natexlab{b}.
\newblock \bibinfo{title}{Link for {GloVe.6B}}.
\newblock
\newblock
\newblock
\shownote{\url{https://nlp.stanford.edu/data/glove.6B.zip}}.


\bibitem[Qin et~al\mbox{.}(2020)]%
        {qin2020ghashing}
\bibfield{author}{\bibinfo{person}{Zongyue Qin}, \bibinfo{person}{Yunsheng
  Bai}, {and} \bibinfo{person}{Yizhou Sun}.} \bibinfo{year}{2020}\natexlab{}.
\newblock \showarticletitle{GHashing: Semantic Graph Hashing for Approximate
  Similarity Search in Graph Databases}. In \bibinfo{booktitle}{\emph{SIGKDD}}.
\newblock


\bibitem[Shu and Nakayama(2018)]%
        {shu2017compressing}
\bibfield{author}{\bibinfo{person}{Raphael Shu} {and} \bibinfo{person}{Hideki
  Nakayama}.} \bibinfo{year}{2018}\natexlab{}.
\newblock \showarticletitle{Compressing Word Embeddings via Deep Compositional
  Code Learning}. In \bibinfo{booktitle}{\emph{ICLR}}.
\newblock


\bibitem[Suzuki and Nagata(2016)]%
        {suzuki2016learning}
\bibfield{author}{\bibinfo{person}{Jun Suzuki} {and} \bibinfo{person}{Masaaki
  Nagata}.} \bibinfo{year}{2016}\natexlab{}.
\newblock \showarticletitle{Learning compact neural word embeddings by
  parameter space sharing}. In \bibinfo{booktitle}{\emph{IJCAI}}.
  \bibinfo{pages}{2046--2052}.
\newblock


\bibitem[Svenstrup et~al\mbox{.}(2017)]%
        {svenstrup2017hash}
\bibfield{author}{\bibinfo{person}{Dan Svenstrup},
  \bibinfo{person}{Jonas~Meinertz Hansen}, {and} \bibinfo{person}{Ole
  Winther}.} \bibinfo{year}{2017}\natexlab{}.
\newblock \showarticletitle{Hash embeddings for efficient word
  representations}.
\newblock \bibinfo{journal}{\emph{arXiv preprint arXiv:1709.03933}}
  (\bibinfo{year}{2017}).
\newblock


\bibitem[Takase and Kobayashi(2020)]%
        {takase2020all}
\bibfield{author}{\bibinfo{person}{Sho Takase} {and} \bibinfo{person}{Sosuke
  Kobayashi}.} \bibinfo{year}{2020}\natexlab{}.
\newblock \showarticletitle{All word embeddings from one embedding}.
\newblock \bibinfo{journal}{\emph{NeurIPS}}  \bibinfo{volume}{33}
  (\bibinfo{year}{2020}), \bibinfo{pages}{3775--3785}.
\newblock


\bibitem[Takase and Okazaki(2019)]%
        {takase2019positional}
\bibfield{author}{\bibinfo{person}{Sho Takase} {and} \bibinfo{person}{Naoaki
  Okazaki}.} \bibinfo{year}{2019}\natexlab{}.
\newblock \showarticletitle{Positional Encoding to Control Output Sequence
  Length}. In \bibinfo{booktitle}{\emph{NAACL-HLT}}.
  \bibinfo{pages}{3999--4004}.
\newblock


\bibitem[Tan et~al\mbox{.}(2020)]%
        {tan2020learning}
\bibfield{author}{\bibinfo{person}{Qiaoyu Tan}, \bibinfo{person}{Ninghao Liu},
  \bibinfo{person}{Xing Zhao}, \bibinfo{person}{Hongxia Yang},
  \bibinfo{person}{Jingren Zhou}, {and} \bibinfo{person}{Xia Hu}.}
  \bibinfo{year}{2020}\natexlab{}.
\newblock \showarticletitle{Learning to hash with graph neural networks for
  recommender systems}. In \bibinfo{booktitle}{\emph{WWW}}.
\newblock


\bibitem[Tang et~al\mbox{.}(2008)]%
        {tang2008arnetminer}
\bibfield{author}{\bibinfo{person}{Jie Tang}, \bibinfo{person}{Jing Zhang},
  \bibinfo{person}{Limin Yao}, \bibinfo{person}{Juanzi Li}, \bibinfo{person}{Li
  Zhang}, {and} \bibinfo{person}{Zhong Su}.} \bibinfo{year}{2008}\natexlab{}.
\newblock \showarticletitle{Arnetminer: extraction and mining of academic
  social networks}. In \bibinfo{booktitle}{\emph{SIGKDD}}.
\newblock


\bibitem[Vaswani et~al\mbox{.}(2017)]%
        {vaswani2017attention}
\bibfield{author}{\bibinfo{person}{Ashish Vaswani}, \bibinfo{person}{Noam
  Shazeer}, \bibinfo{person}{Niki Parmar}, \bibinfo{person}{Jakob Uszkoreit},
  \bibinfo{person}{Llion Jones}, \bibinfo{person}{Aidan~N Gomez},
  \bibinfo{person}{{\L}ukasz Kaiser}, {and} \bibinfo{person}{Illia
  Polosukhin}.} \bibinfo{year}{2017}\natexlab{}.
\newblock \showarticletitle{Attention is all you need}. In
  \bibinfo{booktitle}{\emph{NeurIPS}}.
\newblock


\bibitem[Wang et~al\mbox{.}(2019)]%
        {wang2019neural}
\bibfield{author}{\bibinfo{person}{Xiang Wang}, \bibinfo{person}{Xiangnan He},
  \bibinfo{person}{Meng Wang}, \bibinfo{person}{Fuli Feng}, {and}
  \bibinfo{person}{Tat-Seng Chua}.} \bibinfo{year}{2019}\natexlab{}.
\newblock \showarticletitle{Neural graph collaborative filtering}. In
  \bibinfo{booktitle}{\emph{SIGIR}}. \bibinfo{pages}{165--174}.
\newblock


\bibitem[Wu et~al\mbox{.}(2019)]%
        {wu2019simplifying}
\bibfield{author}{\bibinfo{person}{Felix Wu}, \bibinfo{person}{Amauri Souza},
  \bibinfo{person}{Tianyi Zhang}, \bibinfo{person}{Christopher Fifty},
  \bibinfo{person}{Tao Yu}, {and} \bibinfo{person}{Kilian Weinberger}.}
  \bibinfo{year}{2019}\natexlab{}.
\newblock \showarticletitle{Simplifying graph convolutional networks}. In
  \bibinfo{booktitle}{\emph{ICML}}.
\newblock


\bibitem[Xu et~al\mbox{.}(2018)]%
        {xu2018powerful}
\bibfield{author}{\bibinfo{person}{Keyulu Xu}, \bibinfo{person}{Weihua Hu},
  \bibinfo{person}{Jure Leskovec}, {and} \bibinfo{person}{Stefanie Jegelka}.}
  \bibinfo{year}{2018}\natexlab{}.
\newblock \showarticletitle{How Powerful are Graph Neural Networks?}. In
  \bibinfo{booktitle}{\emph{ICLR}}.
\newblock


\bibitem[Yeh et~al\mbox{.}(2020a)]%
        {yeh2020towards}
\bibfield{author}{\bibinfo{person}{Chin-Chia~Michael Yeh},
  \bibinfo{person}{Dhruv Gelda}, \bibinfo{person}{Zhongfang Zhuang},
  \bibinfo{person}{Yan Zheng}, \bibinfo{person}{Liang Gou}, {and}
  \bibinfo{person}{Wei Zhang}.} \bibinfo{year}{2020}\natexlab{a}.
\newblock \showarticletitle{Towards a flexible embedding learning framework}.
  In \bibinfo{booktitle}{\emph{ICDMW}}. IEEE, \bibinfo{pages}{605--612}.
\newblock


\bibitem[Yeh et~al\mbox{.}(2022)]%
        {sourceurl}
\bibfield{author}{\bibinfo{person}{Chin-Chia~Michael Yeh},
  \bibinfo{person}{Mengting Gu}, \bibinfo{person}{Yan Zheng},
  \bibinfo{person}{Huiyuan Chen}, \bibinfo{person}{Javid Ebrahimi},
  \bibinfo{person}{Zhongfang Zhuang}, \bibinfo{person}{Junpeng Wang},
  \bibinfo{person}{Liang Wang}, {and} \bibinfo{person}{Wei Zhang}.}
  \bibinfo{year}{2022}\natexlab{}.
\newblock \bibinfo{title}{Source Code}.
\newblock
\newblock
\newblock
\shownote{\url{https://www.dropbox.com/s/1mixmhgbg4wiwtd/release.zip}}.


\bibitem[Yeh et~al\mbox{.}(2020b)]%
        {yeh2020merchant}
\bibfield{author}{\bibinfo{person}{Chin-Chia~Michael Yeh},
  \bibinfo{person}{Zhongfang Zhuang}, \bibinfo{person}{Yan Zheng},
  \bibinfo{person}{Liang Wang}, \bibinfo{person}{Junpeng Wang}, {and}
  \bibinfo{person}{Wei Zhang}.} \bibinfo{year}{2020}\natexlab{b}.
\newblock \showarticletitle{Merchant Category Identification Using Credit Card
  Transactions}. In \bibinfo{booktitle}{\emph{Big Data}}. IEEE,
  \bibinfo{pages}{1736--1744}.
\newblock


\bibitem[{Zhou et al.}(2020)]%
        {zhou2020graph}
\bibfield{author}{\bibinfo{person}{{Zhou et al.}}}
  \bibinfo{year}{2020}\natexlab{}.
\newblock \showarticletitle{Graph neural networks: A review of methods and
  applications}.
\newblock \bibinfo{journal}{\emph{AI Open}} (\bibinfo{year}{2020}).
\newblock


\end{thebibliography}

\clearpage
\appendix

\section{Hashing-based Coding Threshold}
\label{app:collision}
\begin{figure}[ht]
\begin{center}
\includegraphics[width=0.9\linewidth]{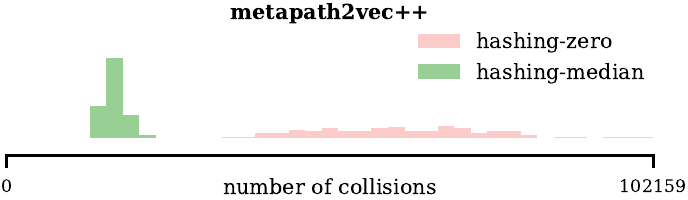}\\
\vspace{0.2cm}
\includegraphics[width=0.9\linewidth]{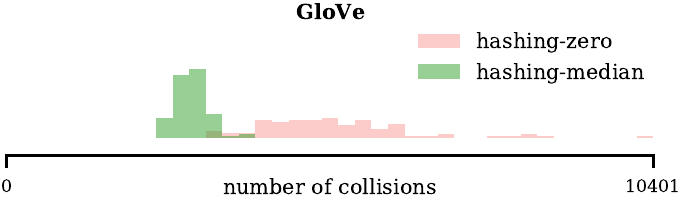}
\end{center}
\caption{
The experiments are performed on \texttt{metapath2vec} and \texttt{GloVe} for 100 times.
The distribution of the 100 outcomes (i.e., number of collisions) for each method is shown in the figure.
The number of collisions is lower for the median threshold compared to the zero threshold.
}
\label{fig:collision_add}
\end{figure}

We have compared the difference between the different choices of thresholds when binarizing the real values into binary codes in Section~\ref{sec:hashing} with an experiment.
Here, we will describe the details of the experiment.
The experiment dataset consists of the first 200,000 pre-trained \texttt{metapath2vec}, \texttt{metapath2vec++} or \texttt{GloVe} embeddings downloaded from the supplemental web page~\cite{dong2017metapath2vec,metapath2vec,pennington2014glove,glove6B}.
The dimension of the pre-trained \texttt{metapath2vec}/\texttt{metapath2vec++} node embeddings is 128.
The dimensionality for the \texttt{GloVe} word embeddings is 300.
Because we repeat the experiments 100 times, we generate 100 seeds to make sure both methods use the same basis to perform random projection as the only difference between the two tested methods should be the threshold.
In each trial, we first use the seed to generated a random matrix~$\mathbf{V} \in \mathbb{R}^{d \times n_{\text{bit}}}$ where $d$ is 128 for node embeddings and 300 for word embeddings.
Next, we project the embedding matrix (i.e., the 200,000 $\times d$  embedding matrix) using $\mathbf{V}$; then, we binarize the result matrix using either zero or the median of each row.
With the binary codes prepared, we count the number of coalitions for each method.
We use 24 bits binary codes in the experiment. 
Once all 100 trials are done, the results are presented with histograms as shown in Figure~\ref{fig:meta_collision_m2v} and Figure~\ref{fig:collision_add}.
The number of collisions is lower for the proposed median threshold compared to the zero threshold baseline.

\section{Pre-trained Embedding}
\label{app:pretrain}

\subsection{Dataset}
\label{app:pretrain_dataset}

\subsubsection{Word embedding}
The pre-trained \texttt{GloVe} word embeddings are downloaded from the web page created by~\cite{pennington2014glove,glove6B}.
The word embeddings are trained using Wikipedia 2014 and Gigaword 5 datasets (total of 6B tokens).

\subsubsection{Word analogy}
We downloaded a list of word analogy pairs from the repository of \texttt{word2vec}~\cite{mikolov2013distributed,word2vec}.
The word analogy pairs are categorized into 14 categories.
The experiment is performed as described by~\cite{mikolov2013distributed}.
Given a word embedding matrix $\mathbf{X}$ and a word analogy pair (e.g., \texttt{Athens:Greece::Bangkok:Thailand}), we first prepare a query vector~$Q$ with $\mathbf{X}[\text{\texttt{Greece}}] - \mathbf{X}[\text{\texttt{Athens}}] + \mathbf{X}[\text{\texttt{Bangkok}}]$.
Next, we use $Q$ to query $\mathbf{X}$ with cosine similarity.
The answer is only considered correct if the most similar word is \texttt{Thailand}.
The performance is measured with accuracy.
We compute the accuracy for each category; then, we report the average of the 14 accuracy values as the performance for word analogy. 

\subsubsection{Word similarity}
Thirteen word similarity datasets are downloaded from the repository of \texttt{MUSE}~\cite{conneau2017word,wordsimilarity,lample2017unsupervised}.
Each dataset consists of a list of paired words and their ground truth similarity scores.
There is a total of 13 datasets.
The experiment is performed as described by~\cite{faruqui2014community}.
First, the cosine similarity between word embeddings for each pair of words in a dataset is computed.
Then, the order based on the cosine similarity is compared with the order based on the ground truth similarity scores.
The comparison of orders is measured with Spearman's rho.
The result Spearman's rhos from the 13 datasets are averaged and reported.

\subsubsection{Node embedding}
The pre-trained \texttt{metapath2vec} embeddings, the pre-trained \texttt{metapath2vec++} embeddings, the association between nodes (i.e., researchers), and the cluster labels (i.e., research area) are downloaded from the web page created by~\cite{dong2017metapath2vec,metapath2vec}.
The node embeddings are trained with AMiner dataset~\cite{tang2008arnetminer}.
There is a total of 246,678 labeled researchers from the downloaded dataset.
Each researcher is assigned one of the eight research areas.
We use $k$-means clustering algorithm~\cite{lloyd1982least} to cluster the embedding associated with each researcher; then, we measure the clustering performance with normalized mutual information.

\subsection{Hyper-parameter Setting}
\label{app:pretrain_hyper}
We use the following hyper-parameter settings for the decoders.
For \texttt{GloVe}, we use $l=3$, $d_c=d_m=512$, $d_e=300$, $c=2$, and $m=128$.
For \texttt{metapath2vec}/\texttt{metapath2vec++}, we use $l=3$, $d_c=d_m=512$, $d_e=128$, $c=2$, and $m=128$.
Note, the decoder design is the same across different coding schemes tested on the same dataset.
We use different $d_e$ for different embeddings because the dimensionality of different pre-trained embeddings is different.
The default hyper-parameter settings for \texttt{AdamW}~\cite{loshchilov2017decoupled} in \texttt{PyTorch}~\cite{paszke2019pytorch} are: $\text{learning rate}=0.001$, $\beta_1=0.9$, $\beta_2=0.999$, and $\text{weight decay}=0.01$.
We train all models for 1,024 epochs with a batch size of 512.

\subsection{Additional Results}
\label{app:pretrain_result}
To understand the relationship between the number of compressed entities and the compression ratio, we construct Table~\ref{tab:compression} to demonstrate how the compression ratio changes as the number of compressed entities are increased.

\begin{table}[ht]
\caption{
Compression ratios for different numbers of compressed entities.
The compression ratios of \texttt{metapath2vec++} are omitted as the compression ratios are the same as \texttt{metapath2vec}.
}
\begin{center}
\footnotesize
\begin{tabular}{l||cccccc}
\# of Entities & 5000 & 10000 & 25000 & 50000 & 100000 & 200000 \\ \hline \hline
\texttt{GloVe} & 2.65 & 5.11  & 11.60 & 20.09 & 31.69  & 44.55  \\
\texttt{metapath2vec} & 1.34 & 2.57  & 5.73  & 9.72  & 14.91  & 20.34 
\end{tabular}
\label{tab:compression}
\end{center}
\end{table}

Aside from the results presented in Figure~\ref{fig:pretrain}, we perform additional experiments to compare the proposed hashing-based coding method with the baseline random coding method under different settings of $c$ and $m$ while varying the number of compressed entities.
The results are presented in Table~\ref{tab:pretrain_add}.
The proposed hashing-based coding method almost always performs better than the baseline random coding method.
The performance gap between the two methods increases as the number of entities compressed by the compression method increases. 
Because the settings of $c$ and $m$ also control the size of the decoder model, $c$ and $m$ affect the compression ratio.
Table~\ref{tab:compression_add} shows the compression ratio under different settings of $c$ and $m$.
Generally, settings with a lower compression ratio have better performance as the potential information loss is less. 
In the experiments, the bit size of the binary code is fixed to 128 bits. 
In other words, both $\{c=256, m=16\}$ and $\{c=2, m=128\}$ use 128 bit binary codes. 
The $c$ and $m$ change the compression ratio by changing the decoder size. 
When using the $\{c=256, m=16\}$ setting, there will be 4,096 vectors total stored in 16 codebooks. 
When using the $\{c=2, m=128\}$ setting, there will be 256 vectors total stored in 2 codebooks. 
Because the $\{c=256, m=16\}$ setting has a larger model (i.e., lower compression ratio), it usually is the setting that outperformed the other in terms of embedding quality. 
To select a suitable setting for $\{c, m\}$, we suggest the users compute the potential memory usage and compression ratio for different settings of $\{c, m\}$, then select the one with the lowest compression ratio while still meets the memory requirement.

\begin{table}[htb]
\caption{
Experiment results on pre-trained embeddings with different settings of $c$ and $m$.
We use \textit{random} to denote the random coding method (i.e., \texttt{ALONE}), and \textit{hashing} to denote the proposed hashing coding method.
}
\begin{center}
\footnotesize
\begin{tabular}{l|ll|l||cccc}
\multirow{2}{*}{}     & \multirow{2}{*}{$c$}   & \multirow{2}{*}{$m$}   & \multirow{2}{*}{Coding Method} & \multicolumn{4}{c}{\# of Entities} \\
                                    &                      &                      &                                & 5000    & 10000  & 50000  & 200000 \\ \hline \hline
\multirow{8}{*}{\rotatebox[origin=c]{90}{\texttt{GloVe} (analogy)}}    & \multirow{2}{*}{2}   & \multirow{2}{*}{128} & random                         & 0.578  & 0.444 & 0.074 & 0.005 \\
                                    &                      &                      & hashing                        & 0.580  & 0.490 & 0.364 & 0.288 \\ \cline{2-8} 
                                    & \multirow{2}{*}{4}   & \multirow{2}{*}{64}  & random                         & 0.593  & 0.460 & 0.095 & 0.007 \\
                                    &                      &                      & hashing                        & 0.601  & 0.487 & 0.320 & 0.294 \\ \cline{2-8} 
                                    & \multirow{2}{*}{16}  & \multirow{2}{*}{32}  & random                         & 0.621  & 0.536 & 0.151 & 0.013 \\
                                    &                      &                      & hashing                        & 0.625  & 0.500 & 0.360 & 0.260 \\ \cline{2-8} 
                                    & \multirow{2}{*}{256} & \multirow{2}{*}{16}  & random                         & 0.671  & 0.653 & 0.426 & 0.084 \\
                                    &                      &                      & hashing                        & 0.668  & 0.669 & 0.471 & 0.314 \\ \hline \hline
\multirow{8}{*}{\rotatebox[origin=c]{90}{\texttt{GloVe} (similarity)}} & \multirow{2}{*}{2}   & \multirow{2}{*}{128} & random                         & 0.544  & 0.517 & 0.371 & 0.106 \\
                                    &                      &                      & hashing                        & 0.544  & 0.539 & 0.526 & 0.411 \\ \cline{2-8} 
                                    & \multirow{2}{*}{4}   & \multirow{2}{*}{64}  & random                         & 0.580  & 0.548 & 0.430 & 0.222 \\
                                    &                      &                      & hashing                        & 0.580  & 0.523 & 0.484 & 0.410 \\ \cline{2-8} 
                                    & \multirow{2}{*}{16}  & \multirow{2}{*}{32}  & random                         & 0.550  & 0.581 & 0.450 & 0.162 \\
                                    &                      &                      & hashing                        & 0.550  & 0.530 & 0.447 & 0.430 \\ \cline{2-8} 
                                    & \multirow{2}{*}{256} & \multirow{2}{*}{16}  & random                         & 0.574  & 0.567 & 0.525 & 0.361 \\
                                    &                      &                      & hashing                        & 0.574  & 0.574 & 0.531 & 0.435 \\ \hline \hline
\multirow{12}{*}{\rotatebox[origin=c]{90}{\texttt{metapath2vec}}}      & \multirow{3}{*}{2}   & \multirow{3}{*}{128} & random                         & 0.773  & 0.764 & 0.723 & 0.603 \\
                                    &                      &                      & hashing/pre-trained          & 0.773  & 0.765 & 0.756 & 0.742 \\
                                    &                      &                      & hashing/graph                & 0.779  & 0.768 & 0.747 & 0.717 \\ \cline{2-8} 
                                    & \multirow{3}{*}{4}   & \multirow{3}{*}{64}  & random                         & 0.772  & 0.769 & 0.727 & 0.627 \\
                                    &                      &                      & hashing/pre-trained          & 0.780  & 0.770 & 0.751 & 0.751 \\
                                    &                      &                      & hashing/graph                & 0.777  & 0.772 & 0.753 & 0.717 \\ \cline{2-8} 
                                    & \multirow{3}{*}{16}  & \multirow{3}{*}{32}  & random                         & 0.776  & 0.772 & 0.737 & 0.669 \\
                                    &                      &                      & hashing/pre-trained          & 0.776  & 0.767 & 0.753 & 0.740 \\
                                    &                      &                      & hashing/graph                & 0.776  & 0.779 & 0.764 & 0.742 \\ \cline{2-8} 
                                    & \multirow{3}{*}{256} & \multirow{3}{*}{16}  & random                         & 0.779  & 0.781 & 0.762 & 0.726 \\
                                    &                      &                      & hashing/pre-trained          & 0.779  & 0.777 & 0.774 & 0.758 \\
                                    &                      &                      & hashing/graph                & 0.781  & 0.780 & 0.760 & 0.749 \\ \hline \hline
\multirow{12}{*}{\rotatebox[origin=c]{90}{\texttt{metapath2vec++}}}    & \multirow{3}{*}{2}   & \multirow{3}{*}{128} & random                         & 0.755  & 0.759 & 0.716 & 0.580 \\
                                    &                      &                      & hashing/pre-trained          & 0.759  & 0.757 & 0.736 & 0.726 \\
                                    &                      &                      & hashing/graph                & 0.754  & 0.750 & 0.734 & 0.701 \\ \cline{2-8} 
                                    & \multirow{3}{*}{4}   & \multirow{3}{*}{64}  & random                         & 0.762  & 0.748 & 0.726 & 0.613 \\
                                    &                      &                      & hashing/pre-trained          & 0.761  & 0.746 & 0.738 & 0.712 \\
                                    &                      &                      & hashing/graph                & 0.759  & 0.753 & 0.740 & 0.703 \\ \cline{2-8} 
                                    & \multirow{3}{*}{16}  & \multirow{3}{*}{32}  & random                         & 0.755  & 0.750 & 0.715 & 0.644 \\
                                    &                      &                      & hashing/pre-trained          & 0.765  & 0.752 & 0.746 & 0.731 \\
                                    &                      &                      & hashing/graph                & 0.761  & 0.756 & 0.742 & 0.727 \\ \cline{2-8} 
                                    & \multirow{3}{*}{256} & \multirow{3}{*}{16}  & random                         & 0.763  & 0.764 & 0.746 & 0.706 \\
                                    &                      &                      & hashing/pre-trained          & 0.760  & 0.766 & 0.750 & 0.743 \\
                                    &                      &                      & hashing/graph                & 0.766  & 0.764 & 0.747 & 0.729
\end{tabular}
\label{tab:pretrain_add}
\end{center}
\end{table}

\begin{table}[htb]
\caption{Compression ratios for different numbers of compressed entities with different settings of $c$ and $m$.
The compression ratios of \texttt{metapath2vec++} are omitted as the ratios are the same as \texttt{metapath2vec}.}
\begin{center}
\footnotesize
\begin{tabular}{l|ll||cccc}
\multirow{2}{*}{Embedding}    & \multirow{2}{*}{$c$} & \multirow{2}{*}{$m$} & \multicolumn{4}{c}{\# of Entities} \\
                              &                    &                    & 5000   & 10000   & 50000  & 200000  \\ \hline \hline
\multirow{4}{*}{\texttt{GloVe}}        & 2                  & 128                & 2.65   & 5.11    & 20.09  & 44.55   \\
                              & 4                  & 64                 & 2.65   & 5.11    & 20.09  & 44.55   \\
                              & 16                 & 32                 & 2.15   & 4.18    & 17.09  & 40.60   \\
                              & 256                & 16                 & 0.59   & 1.18    & 5.53   & 18.11   \\ \hline
\multirow{4}{*}{\texttt{metapath2vec}} & 2                  & 128                & 1.34   & 2.57    & 9.72   & 20.34   \\
                              & 4                  & 64                 & 1.34   & 2.57    & 9.72   & 20.34   \\
                              & 16                 & 32                 & 1.05   & 2.03    & 8.10   & 18.42   \\
                              & 256                & 16                 & 0.26   & 0.52    & 2.44   & 7.94   
\end{tabular}
\label{tab:compression_add}
\end{center}
\end{table}

\section{Node Classification and Link Prediction}
\label{app:gnn_task}

\subsection{Hyper-parameter Setting}
\label{app:gnn_hyper}
We use the following hyper-parameter settings for the decoders: $l=3$, $d_c=d_m=512$, and $d_e=64$.
We use validation data to tune the settings of $c$, the settings of $m$, and the light/full method.
We use the following hyper-parameter settings for the \texttt{GraphSAGE} model: $\text{number of layers}=2$, $\text{number of neurons}=128$, $\text{activation function}=\texttt{ReLU}$, and $\text{number of neighbors}=15$.
These settings are the default hyper-parameter settings from the \texttt{GraphSAGE} implementation~\cite{pytorchgraphsage}.
For \texttt{GCN}~\cite{kipf2016semi}, we use a two-layered structure with a hidden dimension of 128, self-loop, and skip connection.
For \texttt{SGC}~\cite{wu2019simplifying} and \texttt{GIN}~\cite{xu2018powerful}, we also use a two-layered structure with hidden dimension of 128 with the other hyper parameter set to the default values in the PyG library~\cite{fey2019fast}.
We use the following hyper-parameter settings for the \texttt{AdamW} optimizer~\cite{loshchilov2017decoupled}: $\text{learning rate}=0.01$, $\beta_1=0.9$, $\beta_2=0.999$, and $\text{weight decay}=0$.
We train \texttt{GraphSAGE} models for 10 epochs with a batch size of 256 and report the evaluation accuracy from the epoch with the best validation accuracy.
We do not use mini-batches with \texttt{GCN}~\cite{kipf2016semi}, \texttt{SGC}~\cite{wu2019simplifying}, and \texttt{GIN}~\cite{xu2018powerful}; these models are trained for 512 epochs, and the evaluation accuracy from the epoch associated with the best validation accuracy is reported.

\end{document}